\documentclass[utf8]{frontiersSCNS_preprint} % for Science, Engineering and Humanities and Social Sciences articles
\usepackage{url,microtype,subcaption}
\usepackage{lineno}
\usepackage{listings}
\usepackage[onehalfspacing]{setspace}
\usepackage[normalem]{ulem}
\usepackage[]{hyperref}
\hypersetup{
    colorlinks,
    linkcolor={blue!30!black},
    citecolor={blue!30!black},
    urlcolor={white!10!black}
}

\def\namedlabel#1#2{\begingroup
    #2%
    \def\@currentlabel{#2}%
    \phantomsection\label{#1}\endgroup
}

\usepackage{amsmath}
\usepackage{algorithm}
\usepackage[noend]{algpseudocode}

\makeatletter
\def\BState{\State\hskip-\ALG@thistlm}
\makeatother
\usepackage{enumitem}
\def\keyFont{\fontsize{8}{11}\helveticabold }
\def\firstAuthorLast{Eppe {et~al.}} %use et al only if is more than 1 author
\def\Authors{Manfred Eppe\,$^{1,*}$, Phuong D.H. Nguyen\,$^{1}$, and Stefan Wermter\,$^{1}$}
\def\Address{$^{1}$Knowledge Technology Institute, Department of Informatics, Universit\"at Hamburg, Germany
}
\def\corrAuthor{Manfred Eppe}

\def\corrEmail{eppe@informatik.uni-hamburg.de}

\newcommand{\new}[1]{\color{black!50!black}#1\color{black}}

\begin{document}
\onecolumn
\firstpage{1}

% \title[From semantics to execution]{From semantics to execution: Integrating action planning with reinforcement learning for robotic tool use}

\title[From semantics to execution]{From semantics to execution: Integrating action planning with reinforcement learning for robotic causal problem-solving}

\author[\firstAuthorLast ]{\Authors} %This field will be automatically populated
\address{} %This field will be automatically populated
\correspondance{} %This field will be automatically populated

\extraAuth{}% If there are more than 1 corresponding author, comment this line and uncomment the next one.
%\extraAuth{corresponding Author2 \\ Laboratory X2, Institute X2, Department X2, Organization X2, Street X2, City X2 , State XX2 (only USA, Canada and Australia), Zip Code2, X2 Country X2, email2@uni2.edu}

\maketitle

\begin{abstract}
\new{Reinforcement learning is generally accepted to be an appropriate and successful method to learn robot control. } Symbolic action planning is useful to resolve causal dependencies and to break a causally complex problem down into a sequence of simpler high-level actions. A problem with the integration of both approaches is that action planning is based on \emph{discrete high-level action- and state spaces}, whereas reinforcement learning is usually driven by a \emph{continuous reward} function. Recent advances in \new{model-free } reinforcement learning, specifically, universal value function approximators and hindsight experience replay, have focused on goal-independent methods \new{based on \emph{sparse rewards} that are only given at the end of a rollout, and only if the goal has been fully achieved. } In this article, we build on these novel methods to facilitate the integration of action planning with \new{model-free } reinforcement learning. \new{Specifically, the paper demonstrates how the reward-sparsity can serve as a bridge between the high-level and low-level state- and action spaces. }
As a result, we demonstrate that the integrated method is able to solve 
%object manipulation problems that involve tool use and non-trivial causal dependencies 
\new{robotic tasks that involve non-trivial causal dependencies }
under noisy conditions, exploiting both data and knowledge.

\tiny
 \keyFont{ \section{Keywords:} reinforcement learning, hierarchical architecture, planning, robotics, neural networks, causal puzzles} %All article types: you may provide up to 8 keywords; at least 5 are mandatory.
\end{abstract}

\section{Introduction}
\label{sec:intro}
\begin{figure}
    \centering
    \includegraphics[trim=600px 450px 400px 50px,clip,width=0.49\textwidth]{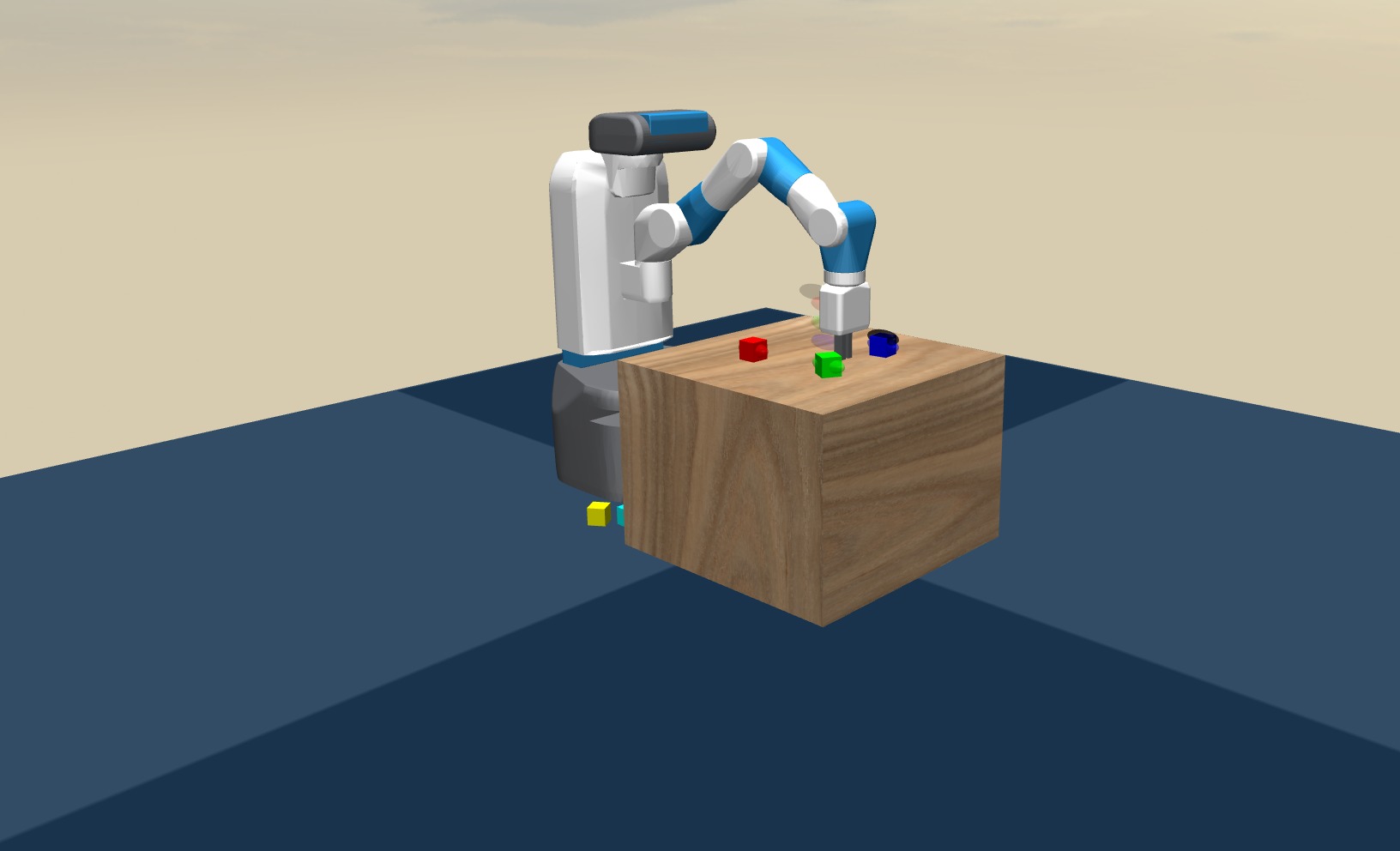}
    \vspace{0.1pt}
    \includegraphics[trim=475px 320px 500px 160px,clip,width=0.49\textwidth]{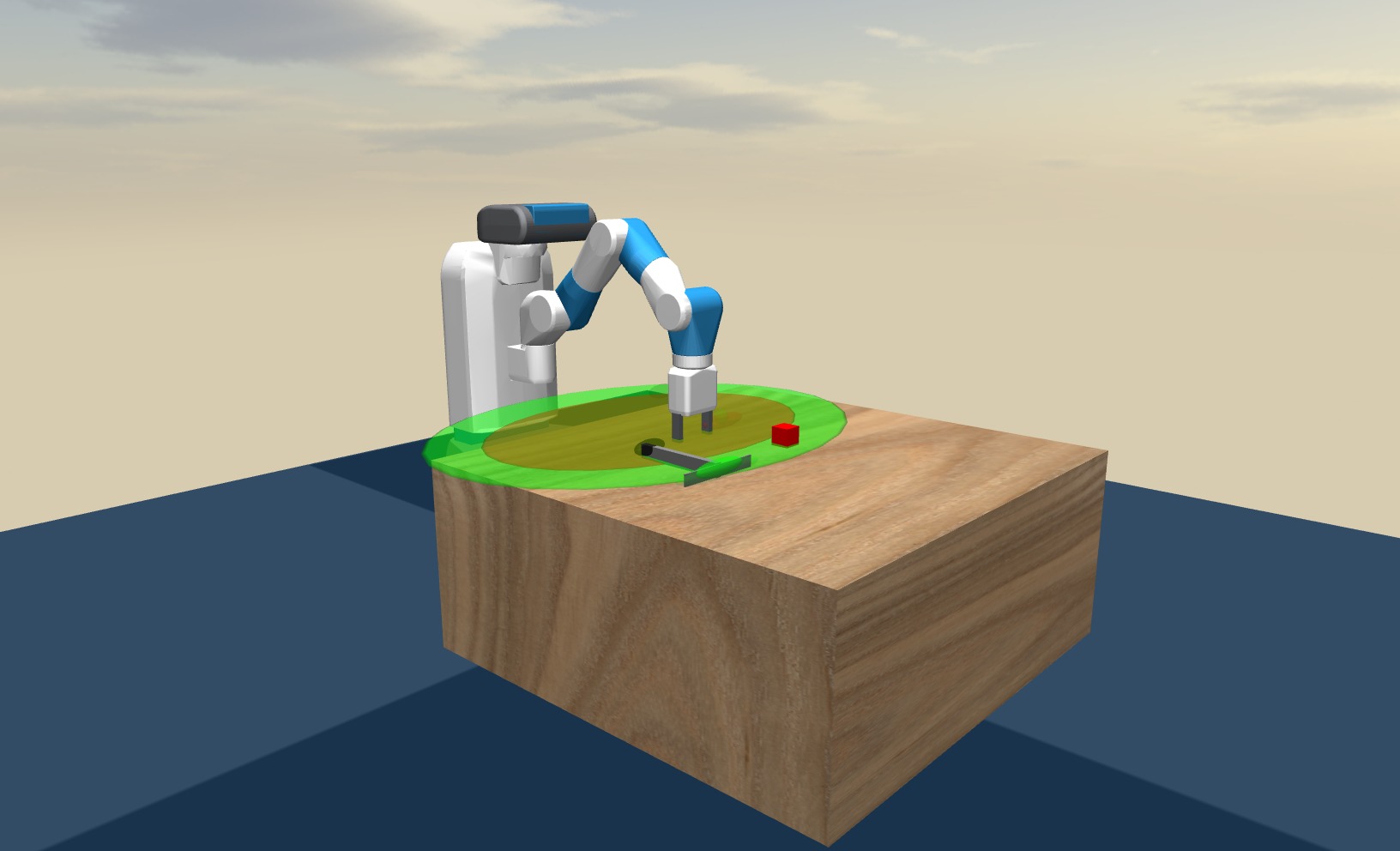}
    \caption{A robot performing two object manipulation tasks. \emph{1. Block-stacking (left):} The gripper must stack three blocks at a random location within the robot's range on the table (indicated by the transparent goal markers behind the gripper). Herein, the robot needs to subdivide the course of actions into separate actions for grasping and placing the individual blocks. 
     \emph{2. Tool use (right):} The red block is out of the gripper's range (indicated by the dark brown ellipsoid), so that solving the task of moving the block to a target location requires the robot to break the problem down into a sequence of high-level actions that involve grasping the rake, moving the rake towards the block and pulling the rake. 
     }
    \label{fig:tool_use}
\end{figure}

\emph{How can one realize robots that reason about complex physical object manipulation problems, and how can we integrate this reasoning with the noisy sensorimotor machinery that executes the required actions in a continuous low-level action space?}
To address these research questions, we consider reinforcement learning (RL) as it is a successful method to facilitate low-level robot control \citep{Deisenroth2011_PILCO}. It is well known that non-hierarchical reinforcement-learning architectures fail in situations involving non-trivial causal dependencies that require the reasoning over an extended time horizon \citep{Mnih2015}. 
For example, the robot in \autoref{fig:tool_use} (right) needs to first grasp the rake before it can be used to drag the block to a target location. Such a problem is hard to solve by RL-based low-level motion planning without any high-level method that subdivides the problem into smaller sub-tasks. 

To this end, recent research has developed hierarchical and model-based reinforcement learning methods to tackle problems that require reasoning over a long time horizon, as demanded in domains like robotic tool use, block-stacking \citep{Deisenroth2011_PILCO}, and computer games \citep{Aytar2018_DQN_Atari_Youtube,Pohlen2018_Atari_Consistent}. 
Yet, the problem of realizing an agent that learns to solve open-domain continuous space causal puzzles from scratch, without learning from demonstration or other data sources, remains unsolved.
The existing learning-based approaches are either very constrained (e.g., \citep{Deisenroth2011_PILCO}), or they have been applied only to low-dimensional non-noisy control problems that do not involve complex causal dependencies (e.g., \citep{Bacon2016_OptionCritic_HRL,Levy2019_Hierarchical}),  or they build on learning from demonstration \citep{Aytar2018_DQN_Atari_Youtube}.

A complementary method to address complex causal dependencies is to use pre-specified semantic domain knowledge, e.g., in the form of an action planning domain description \citep{Mcdermott1998_PDDL}. With an appropriate problem description, a planner can provide a sequence of solvable sub-tasks in a discrete high-level action space. 
However, the problem with this semantic and symbolic task planning approach is that the high-level actions generated by the planner require the grounding in a low-level motion execution layer to consider the context of the current low-level state. 
For example, executing a high-level robotic action \texttt{move\_object\_to\_target} requires precise information (e.g., the location and the shape) of the object to move it to the target location along a context-specific path.
This grounding problem consists of two sub-problems \ref{itm:P1} and \ref{itm:P2} that we tackle in this article. 

\begin{enumerate}[leftmargin=2.5em,labelsep=1em,label=\textbf{P.\arabic*}]
    \item \label{itm:P1} \textit{The first grounding sub-problem is to map the discrete symbolic action space to context-specific subgoals in the continuous state space.} For instance, \texttt{move\_object\_to\_target} needs to be associated with a continuous sub-goal that specifies the desired metric target coordinates of the object.
    \item \label{itm:P2}  \textit{The second grounding sub-problem is to map the subgoals to low-level context-specific action trajectories.} For instance, the low-level trajectory for \texttt{move\_object\_to\_target} is specific to the continuous start- and target location of the object, and to potential obstacles between start and target.
\end{enumerate}

\new{Both problems are currently being recognized by the state of the art in combined task and motion planning (e.g., \cite{Toussaint2018_DiffPhysics_ToolUse}), and, from a broader perspective, also in the field of state representation learning (e.g., \cite{Lesort2018statel,Doncieux2018}). However, to the best of our knowledge, there exist currently no satisfying and scalable solutions to these problems that have been demonstrated in the robotic application domain with continuous state- and action representations (see \autoref{sec:sota:summary}).} 
In this research, we address \ref{itm:P1} by providing a simple, yet principled, formalism to map the propositional high-level state space to continuous subgoals (\autoref{sec:method:abstraction_grounding}). We address \ref{itm:P2} by integrating this formalism with goal-independent reinforcement learning based on sparse rewards (\autoref{sec:method:rl}). 
% \new{To some extent, the utilized approach of goal-independent reinforcement learning is a special case of the multiple-goal exploration process~\citep{Forestier2017IntrinsicallyLearning} ... }

Existing approaches that integrate action planning with reinforcement learning have not been able to map subgoals to low-level motion trajectories for realistic continuous-space robotic applications~\citep{Grounds2005_Plan-Q,Ma2009_PolicySearch_Planning} because they rely on a continuous dense reward signal that is proportional to manually defined metrics that estimate how well a problem has been solved~\citep{Ng1999_RewardShaping}. 
The manual definition of such metrics, also known as \emph{reward shaping}, is a non-trivial problem itself because the semantic distance to a continuous goal is often not proportional to the metric distance.

Recently, so-called \emph{universal value function approximators} (UVFAs)~\citep{Schaul2015_UVFA} in combination with \emph{hindsight experience replay} (HER)~\citep{Andrychowicz2017HindsightReplay} and neural actor-critic reinforcement learning methods \citep{Lillicrap2016_DDPG} have been proposed to alleviate this issue. HER realizes an efficient off-policy algorithm that allows for non-continuous sparse rewards without relying on reward shaping. Specifically, HER treats action trajectories as successful that do not achieve the desired specified goal, by pretending in hindsight that the achieved state was the desired goal state.
Our research builds on this method because the sparse subgoal-specific rewards allow us to decouple the reward mechanism from the high-level action planning. 

% \pagebreak

This approach enables us to address the following \textbf{central hypotheses:}
\new{
\begin{enumerate}[leftmargin=2.5em,labelsep=1em,label=H.\arabic*]
    \item \label{itm:H1} \textit{We hypothesize that model-free reinforcement learning with universal value function approximators (UVFAs) and hindsight experience replay (HER) is appropriate to learn the grounding of a discrete symbolic action space in continuous action trajectories. We measure the appropriateness by comparing the resulting hybrid discrete/continuous architecture with continuous hierarchical reinforcement learning (HRL). We consider our approach to be appropriate if it is better capable of learning to solve causal object-manipulation puzzles that involve tool use and causal chains of non-trivial length that HRL.}
    \item \label{itm:H2} \textit{We hypothesize that the approach is robust enough to handle a realistic amount of perceptual noise. We consider the approach to be robust to noise if there is no significant performance drop when moderate noise, e.g., 1-2\% of the observational range, is added to the agent's state representation.}
\end{enumerate}
}
\new{We address these hypotheses by applying our method to three simulated robotic environments that are based on the OpenAI Gym framework. } For these environments, we provide manually defined action planning domain descriptions and combine a planner with a \new{model-free } reinforcement learner to learn the grounding of high-level action descriptions in low-level trajectories. 

\textbf{Our research contribution} is \textit{a principled method and proof-of-concept to ground high-level semantic actions in low-level sensorimotor motion trajectories and to integrate \new{model-free } reinforcement learning with symbolic action planning.}
\new{The novelty of this research is to use UVFAs and HER to decouple the reward mechanism from the high-level propositional subgoal representations provided by the action planner: Instead of defining an individual reward function for each predicate, our approach allows for a single simple threshold-based sparse reward function that is the same for all predicates.}

% Herein, we do not focus on achieving good benchmark results in comparison with other approaches. An empirical comparison is also not straight-forward, because most existing approaches are either limited to discrete action- and state spaces or they rely on other non-trivial forms of domain-knowledge. \textbf{Our research goal} is rather \emph{to provide a proof-of-concept and a baseline for the integration of action planning with reinforcement learning in continuous domains that involve complex causal dependencies.
\textbf{Our research goal} is \emph{to provide a proof-of-concept and a baseline for the integration of action planning with reinforcement learning in continuous domains that involve complex causal dependencies.}

The remainder of the article is organized as follows. In \autoref{sec:sota} we investigate the state of the art in task and motion planning, hierarchical learning and the integration of planning with learning. We identify the problem of grounding high-level actions in low-level trajectories as a critical issue for robots to solve causal puzzles. 
We present our method and the underlying background in \autoref{sec:method}. We describe the realization of our experiments in \autoref{sec:experiments} and show the experimental results in \autoref{sec:results} before we discuss and align our findings with the hypotheses in \autoref{sec:discussion}. We conclude in \autoref{sec:conclusion}.

\section{State of the art}
\label{sec:sota}
Our work is related to robotic task and motion planning, but it also addresses plan execution. Therefore, it is also related to hierarchical learning algorithms and the integration of learning with planning. 

\subsection{Combined task and motion planning}
\label{sec:sota:tamp}
The field of combined task and motion planning (TAMP) investigates methods to integrate low-level motion planning with high-level task planning. The field aims at solving physical puzzles and problems that are too complex to solve with motion planning alone, often inspired by smart animal behavior \citep{Toussaint2018_DiffPhysics_ToolUse}. For example, crows are able to perform a sequence of high-level actions, using tools like sticks, hooks or strings, to solve a puzzle that eventually leads to a reward \citep{Taylor2009}. 
A set of related benchmark problems has recently been proposed by \citet{Lagriffoul2018_TAMP_Benchmarks}. However, since TAMP focuses primarily on the planning aspects and not necessarily on the online action execution, the benchmark environments do not consider a physical action execution layer. 

\citet{Toussaint2018_DiffPhysics_ToolUse} formulate the TAMP problem as an inverted differentiable physics simulator. The authors consider the local optima of the possible physical interactions by extending mixed-integer programs (MIP)~\citep{Deits2014_MIP} to first-order logic. The authors define physical interactions as action primitives that are grounded in contact switches. The space of possible interactions is restricted to consider only those interactions that are useful for the specific problem to solve. These interactions are formulated based on a fixed set of predicates and action primitives in the domain of robotic tool use and object manipulation. 
However, the authors provide only a theoretical framework for planning, and they do not consider the physical execution of actions. Therefore, an empirical evaluation to measure the actual performance of their framework, considering also real-world issues like sensorimotor noise, is not possible. 

Other TAMP approaches include the work by \citet{Alili2010_SymGeoReasoning} and \citet{DeSilva2013_HTN_GeoPlanning} who both combine a hierarchical symbolic reasoner with a geometrical reasoner to plan human-robot handovers of objects. Both approaches consider only the planning, not the actual execution of the actions. The authors do not provide an empirical evaluation in a simulated or real environment. 
\citet{Srivastava2014_TAMP_Planner_Interface} also consider action execution and address the problem of grounding high-level tasks in low-level motion trajectories by proposing a planner-independent interface layer for TAMP that builds on symbolic references to continuous values. Specifically, they propose to define symbolic actions and predicates such that they refer to certain objects and their poses. They leave it to the low-level motion planner to resolve the references. Their approach scales well on the planning level in very cluttered scenes, i.e., the authors demonstrate that the planning approach can solve problems with 40 objects. The authors also present a physical demonstrator using a PR2 robot, but they do not provide a principled empirical evaluation to measure the success of the action execution under realistic or simulated physical conditions. \citet{Wang2018_ModelLearningSampling_TAMP} also consider action execution, though only in a simple 2D environment without realistic physics. Their focus is on solving long-horizon task planning problems that involve sequences of 10 or more action primitives. To this end, the authors present a method that learns the conditions and effects of high-level action operators in a kitchen environment. 

A realistic model that also considers physical execution has been proposed by \citet{Leidner2018}. The authors build on geometric models and a particle distribution model to plan goal-oriented wiping motions. Their architecture involves low-level and high-level inference and a physical robotic demonstrator. However, the authors build on geometric modeling and knowledge only, without providing a learning component. Noisy sensing is also not addressed in their work.

\subsection{Hierarchical learning-based approaches}
\label{sec:sota:hrl}
Most TAMP approaches consider the  planning as an offline process given geometrical, algebraic or logical domain models. This, however, does not necessarily imply the consideration of action execution. 
The consideration of only the planning problem under idealized conditions is not appropriate in practical robotic applications that often suffer from sensorimotor noise. 
To this end, researchers have investigated hierarchical learning-based approaches that differ conceptually from our work because they build on data instead of domain-knowledge to realize the high-level control framework.

For example, \cite{Levy2019_Hierarchical} and \cite{Nachum2018_HIRO} both consider ant-maze problems in a continuous state- and action space. 
The challenge of these problems lies in coordinating the low-level walking behavior of the ant-like agent with high-level navigation. 
\new{However, these approaches do not appreciate that different levels of the problem-solving process require different representational abstractions of states and actions. For example, in our approach, the planner operates on propositional state descriptions like ``object 1 on top of object 2'' and generates high-level conceptual actions like ``move gripper to object''. In those HRL approaches, the high-level state- and action representations are within the same state-and action space as the low-level representations. This leads to larger continuous problem spaces.}

Other existing hierarchical learning-based approaches are limited to discrete action- or state spaces on all hierarchical layers. For example, \citet{Kulkarni2016} present the h-DQN framework to integrate hierarchical action-value functions with goal-driven intrinsically motivated deep RL. Here, the bottom-level reward needs to be hand-crafted using prior knowledge of applications. \citet{Vezhnevets2017_Feudal} introduce the FeUdal Networks (FuNs), a two-layer hierarchical agent. The authors train subgoal embeddings to achieve a significant performance in the context of Atari games with a discrete action space. Another example of an approach that builds on discrete actions is the option-critic architecture by \citet{Bacon2016_OptionCritic_HRL}. 
Their method extends gradient computations of intra-option policies and termination functions to enable learning options that maximize the expected return within the options framework, proposed by \citet{sutton_between_1999}. The authors apply and evaluate their framework in the Atari gaming domain.

\subsection{Integrating learning and planning}
\label{sec:sota:learning_planning}
There exist several robot deliberation approaches that exploit domain knowledge to deliberate robotic behavior and to perform reasoning (e.g., \cite{Eppe2013_LPNMR,Rockel2013}). The following examples from contemporary research extend the knowledge-based robotic control approach and combine it with reinforcement learning:

The Dyna Architecture~\citep{Sutton1991_dyna} and its derived methods, e.g., Dyna-Q~\citep{sutton_integrated_1990}, queue-Dyna~\citep{peng_efficient_1993}, RTP-Q~\citep{zhao_rtp-q:_1999}, aim to speed up the learning procedure of the agent by unifying reinforcement learning and incremental planning within a single framework. While the RL component aims to construct the action model as well as to improve the value function and policy directly through real experiences from environment interaction, the planning component updates the value function with simulated experiences collected from the action model. 
The authors show that instead of selecting uniformly experienced state-action pairs during the planning, it is much more efficient to focus on pairs leading to the goal state (or nearby states) because these cause larger changes in value function. This is the main idea of the prioritized sweeping method~\citep{Moore1993PrioritizedTime} and derivatives \citep{Andre1998GeneralizedSweeping}.
The methods based on Dyna and prioritized sweeping have neither been demonstrated to address sparse rewards nor do they consider mappings between discrete high-level actions and states and their low-level counter parts.

\cite{Ma2009_PolicySearch_Planning} present the policy search planning method, in which they extend the policy search GPOMDP~\citep{Baxter2001Infinite-horizonEstimation} towards the multi-agent domain of robotic soccer. Herein, they map symbolic plans to policies using an expert knowledge database. The approach does not consider tool use or similar causally complex problems. 
A similar restriction pertains to PLANQ-learning framework \citep{Grounds2005_Plan-Q}: The authors combine a Q-learner with a high-level STRIPS planner~\citep{Fikes1972_STRIPS}, where the symbolic planner shapes the reward function to guide the learners to the desired policy. First, the planner generates a sequence of operators to solve the problem from the problem description. Then, each of these operators is learned successively by the corresponding Q-learner. This discrete learning approach, however, has not been demonstrated to be applicable beyond toy problems, such as the grid world domain that the authors utilize for demonstrations in their paper. 
\citet{Yamamoto2018_Learning_AbductivePlanning} propose a hierarchical architecture that uses a high-level abduction-based planner to generate subgoals for the low-level on-policy reinforcement learning component, which employed the proximal policy optimization (PPO) algorithm~\citep{Schulman2017ProximalAlgorithms}. This approach requires the introduction of an additional task-specific evaluation function, alongside the basic evaluation function of the abduction model to allow the planner to provide the learner with the intrinsic rewards, similar to~\citep{Kulkarni2016}.
The evaluation is only conducted in a grid-based virtual world where an agent has to pick up materials, craft objects and reach a goal position. 

A very interesting approach has been presented by \citet{Ugur2015}. The authors integrate learning with planning, but instead of manually defining the planning domain description they learn it from observations. To this end, they perform clustering mechanisms to categorize object affordances and high-level effects of actions, which are immediately employed in a planning domain description. The authors demonstrate their work on a physical robot. In contrast to our work, however, the authors focus mostly on the high-level inference and not on the robustness that low-level reinforcement-based architectures provide.

\new{
\subsection{Summary of the state of the art}
\label{sec:sota:summary}
% \end{enumerate}

% TODO: (Phuong) Summarize that existing approaches either perform no conceptual abstraction: i) they operate only on the low-level state representations (HRL), or they perform abstraction, but either iia)  using manually defined mapping between high-level and low-level (TAMP, Toussaint2018), or, iib) they learn the mappings but require separate reward functions for each high-level representational element \cite{?}.

The main weaknesses of the related state of the art that we address in this research are the following:
TAMP approaches (\autoref{sec:sota:tamp}) mainly focus on the planning aspect. Whereas they do not consider the physical execution of the planned actions or only evaluate the plan execution utilizing manually defined mapping between high-level (symbolic) and low-level (continuous value). These approaches require domain knowledge and model of robots and the environment to specify and execute the task and motion plans, which may suffer from noisy sensing conditions. On the contrary, hierarchical learning-based approaches (\autoref{sec:sota:hrl}) propose to learn both high-level and low-level from data, but mostly focus on solving problems with discrete action space, and they require internal hand-crafted reward functions. Methods with continuous action space like~\citep{Levy2019_Hierarchical, Nachum2019_HIRO} only consider setups without \new{representational abstractions between the different hierarchical layers. } Mixed approaches (\autoref{sec:sota:learning_planning}) that integrate learning and planning have similar disadvantages as the two other groups. In particular the lack of principled approaches to realize the mapping between discrete and continual spaces, the manual shaping of reward functions, and the lack of approaches that have been demonstrated and applied in a continuous-space realistic environment.

% TODO (Manfred) Align this summary with the extended problem description of P.1. and P.2.

}
\section{Integrating reinforcement learning with action planning}
\label{sec:method}

\begin{figure}[h]
    \centering
    \includegraphics[width=\textwidth]{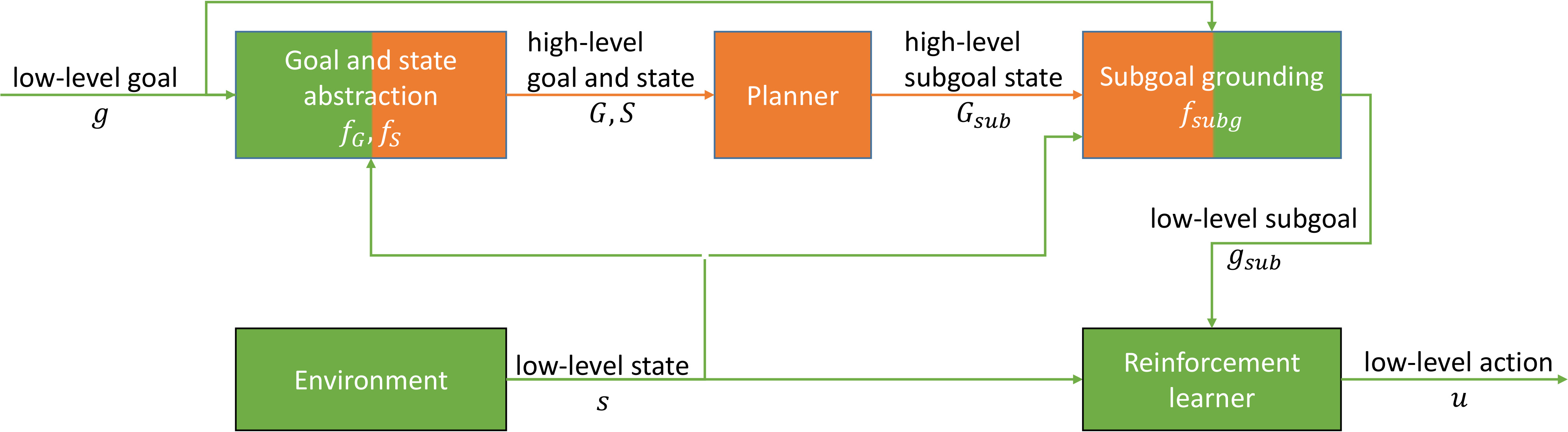}
    \caption{Our proposed integration model. Low-level motion planning elements are indicated in green and high-level elements in orange color. The abstraction functions $f_S,f_G$ map the low-level state and goal representations $s,g$ to high-level state and goal representations $S,G$. These are given as input to the planner to compute a high-level subgoal state $G_{sub}$. The subgoal grounding function $f_{subg}$ maps $G_{sub}$ to a low-level subgoal $g_{sub}$ under consideration of the context provided by the current low-level state $s$ and the low-level goal $g$. The reinforcement learner learns to produce a low-level motion plan that consists of actions $u$ based on the low-level subgoal $g_{sub}$ and the low-level state $s$. }
    \label{fig:architecture}
\end{figure}
To tackle the problems \ref{itm:P1} and \ref{itm:P2}, and to address the research goal of grounding high-level actions in low-level control trajectories, we propose the architecture depicted in \autoref{fig:architecture}. 
\new{
The novelty of the architecture with respect to the state of the art is its ability to learn to achieve subgoals that are provided in an abstract symbolic high-level representation by using a single universal sparse reward function that is appropriate for all discrete high-level goal definitions. 
This involves i) the grounding of the high-level representations to low-level subgoals (\autoref{sec:method:abstraction_grounding}, Algorithm \autoref{alg:fsubg}), and ii) the formalization of the abstraction of the low-level space to the high-level space (\autoref{sec:method:abstraction_grounding}, \autoref{eq:p2bool}), iii) a UVFA- and HER-based reinforcement learner to achieve the low-level subgoals (\autoref{sec:method:rl}), and, iv) the integration of an action planner with the reinforcement learning using the abstraction- and grounding mechanisms (\autoref{sec:method:integration}).
}

\new{The resulting architecture is able to acquire a large repertoire of skills, similar to a multiple-goal exploration process~\citep{Forestier2017IntrinsicallyLearning}. However, instead of sampling and evaluating (sub-)goals through intrinsic rewards, the subgoals within our architecture are generated by the planner.}

\subsection{Abstraction and grounding of states and goals}
\label{sec:method:abstraction_grounding}
Our abstraction and grounding mechanism tackles the research problem \ref{itm:P1}, i.e., the mapping from high-level actions to low-level subgoals. 
STRIPS-based action descriptions are defined in terms of state changes based on predicates for pre- and postconditions. To appreciate that the state change is determined by the postcondition predicates, and not by the actions themselves, it is more succinct to define subgoals in terms of postcondition predicates because multiple actions may involve the same postconditions. Therefore, we define a grounding function for subgoals $f_{subg}$. The function is based on predicates instead of actions to avoid redundancy and to minimize the hand-engineering of domain models and background knowledge. 

To abstract from low-level perception to high-level world states, we define abstraction functions $f_S, f_G$. These functions do not require any additional background knowledge because they fully exploit the definition of $f_{subg}$. 
In our formalization of the abstraction and grounding, we consider the following conventions and assumptions \ref{conv:full_obs}-\ref{conv:goal_idx}:
\begin{enumerate}[leftmargin=2.5em,labelsep=1em,label=C.\arabic*]
    \item\label{conv:full_obs} The low-level environment state space is fully observable, but observations may be noisy. We represent low-level environment states with finite-dimensional vectors $s$. 
    \new{To abstract away from visual preprocessing issues and to focus on the main research questions, we adapt the state representations commonly used in deep reinforcement learning literature \citep{Levy2019_Hierarchical,Andrychowicz2017HindsightReplay}, i.e., states are constituted by the locations, velocities, and rotations of objects (including the robot itself) in the environment.}
    \item The low-level action space is determined by continuous finite-dimensional vectors $u$.
    For example, the \new{robotic manipulation } experiments described in this paper consider a four-dimensional action space that consists of the normalized relative three-dimensional movement of the robot's gripper in Cartesian coordinates plus a scalar value to represent the opening angle of the gripper's fingers.
    \item Each predicate of the planning domain description determines a Boolean property of one object in the environment. The set of all predicates is denoted as $\mathcal{P} = \{p_1,\cdots,p_n\}$. The high-level world state $S$ is defined as the conjunction of all positive or negated predicates. 
    \item The environment configuration is fully determined by a set of objects whose properties can be described by the set of high-level predicates $\mathcal{P}$. Each predicate $p \in \mathcal{P}$ can be mapped to a continuous finite-dimensional low-level substate vector $s_p$. For example, the location property and the velocity property of an object in Cartesian space are both fully determined by a three-dimensional continuous vector.
    \item\label{conv:sequence_pred_overlap} For each predicate $p$, there exists a sequence of indices $p^{idx}$ that determines the indices of the low-level environment state vector $s$ that determines the property described by $p$. For example, given that $p$ refers to the object being at a specific location, and given that the first three values of $s$ determine the Cartesian location of an object, we have that $p^{idx} = [0,1,2]$. 
    A requirement is that the indices of the individual predicates must not overlap, i.e., abusing set notation: $p_1^{idx} \cap p_2^{idx} = \emptyset$ (see Algorithm \ref{alg:fsubg} for details). 
    \item The high-level action space consists of a set of grounded STRIPS operators \citep{Fikes1972_STRIPS} $a$ that are determined by a conjunction of precondition literals and a conjunction of effect literals.
    \item\label{conv:goal_idx} A low-level goal $g$ is the subset of the low-level state $s$, indicated by the indices $g^{idx}$, i.e.,  $g = s[g^{idx}]$. For example, consider that the low-level state $s$ is a six-dimensional vector where the first three elements represent the location and the last three elements represent the velocity of an object in Cartesian space. Then, given that $g^{idx} = [0,1,2]$, we have that $g = s[g^{idx}]$ refers to the location of the object.
\end{enumerate}

\subsubsection{Mapping predicates to low-level subgoals}
Abstracting from observations to high-level predicate specifications is achieved by mapping the low-level state space to a high-level conceptual space. This is realized with a set of functions $f^p_{subg}$ %(\ref{eq:p2subg}) 
that we define manually for each predicate $p$. For a given predicate $p$, the function $f^p_{subg}$ generates the low-level substate $s_p$ that determines $p$, based on the current state $s$ and the goal $g$:

\begin{equation}
\label{eq:p2subg}
\begin{aligned}
    f^{p}_{subg}(s,g) = s_p
\end{aligned}
\end{equation}
\new{
To illustrate how $f^{p}_{subg}$ can be implemented, consider the following two examples from a block-stacking task:
\begin{enumerate}[leftmargin=2.5em,labelsep=1em]
    \item Consider a predicate $\texttt{(at\_target o1)}$ which indicates whether an object is at a given goal location on the surface of a table. Then the respective function $f^{\texttt{(at\_target o1)}}_{subg}(s,g)$ can be implemented as follows:

\begin{equation}
    f^{\texttt{(at\_target o1)}}_{subg}(s,g) = [g[0],g[1],g[2]]
\end{equation}
In this case, the function extracts the respective target coordinates for the object \texttt{o1} directly from $g$ and does not require any information from $s$. 
\item Consider further a predicate $\texttt{(on o2 o1)}$, which is true if an object \texttt{o2} is placed on top of another object \texttt{o1}. Given that the Cartesian location of \texttt{o1} is defined by the first three elements of the state vector $s$, one can define the following subgoal function:
\begin{equation}
    f^{\texttt{(on o2 o1)}}_{subg}(s,g) = [s[0],s[1],s[2]+h_{obj}] \text{, where $h_{obj}$ denotes the height of an object}
\end{equation}
Here, the target coordinates of the object \texttt{o2} are computed by considering the current coordinates of \texttt{o1}, i.e., the first three values of $s$, and by adding a constant for the object height to the third (vertical axis) value.
\end{enumerate}
}

\subsubsection{Grounding high-level representations in low-level subgoals}
The function $f_{subg}$ that maps the high-level subgoal state $G_{sub}$ in the context of $s,g$ to a low-level subgoal $g_{sub}$ builds on Eq.~(\ref{eq:p2subg}), as described with the following Algorithm \ref{alg:fsubg}:

\begin{algorithm}
\caption{Mapping propositional state representations to continuous state representations}\label{alg:fsubg}
\begin{algorithmic}[1]
\Function{$f_{subg}(G_{subg},s,g)$}{}
\State $s_{subg} \gets s$
\State {subgoal\_changed} $\gets$ True
\While {subgoal\_changed}
\State $s_{subg}^{last} \gets s_{subg}$
\For {$p \in \mathcal{P}$} \Comment{For each predicate in high-level goal state, set low-level subgoal indices $p^{idx}$}
\State $s_{subg}[p^{idx}] \gets s_p$, where $s_p = f^{p}_{subg}(s_{subg},g)$ 
\EndFor
\State {subgoal\_changed} $\gets s_{subg}^{last} \neq s_{subg}$
\EndWhile
\EndFunction
\end{algorithmic}
\end{algorithm}

The while loop is necessary to prevent the situation where the application of $f^{p}_{subg}$ (in line 7, Algorithm \autoref{alg:fsubg}) changes $s_{subg}$ in a manner that affects a previous predicate subgoal function. For example, consider the two predicates $\texttt{(at\_target o1)}$ and $\texttt{(on o2 o1)}$. The predicate \texttt{(at\_target o1)} determines the Cartesian location of \texttt{o1}, and \texttt{(on o2 o1)} depends on these coordinates. Therefore, it may happen that $f^{\texttt{(at\_target o1)}}_{subg}=[x,y,z]$ causes the first three elements of $s_{subg}$ to be $x,y,z$. However, $f^{\texttt{(on o2 o1)}}_{subg}=[x',y',z']$ depends on these $x,y,z$ to determine the $x',y',z'$ that encode the location of o2 in $s_{subg}$. The while loop assures that $f^{\texttt{(on o2 o1)}}_{subg}$ is applied at least once after $f^{\texttt{(at\_target o1)}}_{subg}$ to consider this dependency. This assures that all dependencies between the elements of $s_{subg}$ are resolved. 

To guarantee the termination of the Algorithm \autoref{alg:fsubg}, i.e., to avoid that the alternating changes of $s_{subg}$ cause an infinite loop, the indices $p^{idx}$ must be constrained in such a way that they do not overlap (see assumption \ref{conv:sequence_pred_overlap}).

\subsubsection{Abstracting from low-level state- and goal representations to propositional statements}
To realize the abstraction from low-level to high-level representations, we define a set of functions in the form of \autoref{eq:p2bool}. Specifically, for each predicate $p$, we define the following function $f_{p}$ maps the current low-level state and the low-level goal to the predicates' truth values.
\begin{equation}
\label{eq:p2bool}
\begin{aligned}
    f_{p}(p,s) &= \textit{diff}(s[p^{idx}],s_p)  < \epsilon, \text{with } s_p = f^{p}_{subg}(s,g)
\end{aligned}
\end{equation}

\autoref{eq:p2bool} examines whether the subgoal that corresponds to a specific predicate is true, given the current observed state $s$ and a threshold value for the coordinates $\epsilon$. In this article, but without any loss of generality, we assume that each predicate is determined by three coordinates.  \autoref{eq:p2bool} computes the difference between these coordinates given the current state, and the  coordinates determined by $f^p_{subg}$ as their Euclidean distance. 
For example, $f^{\text{\texttt{(at\_target o1)}}}_{subg}$ may define target coordinates for \texttt{o1}, and \autoref{eq:p2bool} considers the distance between the target coordinates and the current coordinates.

\subsection{Generation of adaptive low-level action trajectories}
\label{sec:method:rl}
To address the research problem \ref{itm:P2}, i.e, the grounding of actions and subgoals in low-level action trajectories, we consider continuous goal-independent reinforcement learning approaches \citep{Lillicrap2016_DDPG,Schaul2015_UVFA}.
Most reinforcement learning approaches build on manually defined reward functions based on a metric that is specific to a single global goal, such as the body height, posture and forward speed of a robot that learns to walk \citep{Schulman2015a}.  
Goal-independent reinforcement learning settings do not require such reward shaping (c.f. \citep{Ng1999_RewardShaping}), as they allow one to parameterize learned policies and value functions with goals. 
We employ the actor-critic deep deterministic policy gradient (DDPG) \citep{Lillicrap2016_DDPG} approach in combination with hindsight experience replay (HER) \citep{Andrychowicz2017HindsightReplay} to realize the goal-independent RL \new{for the continuous control } part in our framework.  
Using the HER technique with off-policy reinforcement learning algorithms (like DDPG) increases the efficiency of sampling for our approach since HER stores not only the experienced episode with the original goal ($episode \gets (s_t, u_t,s_{t+1},g)$) in the replay buffer. It also stores modified versions of an episode where the goal is retrospectively set to a state that has been achieved during the episode, i.e., $episode' \gets (s_t, u_t,s_{t+1},g')$ with $g' = s_{t'}[g^{idx}]$ for some $t'>t$. 

To realize this actor-critic architecture, we provide one fully connected neural network for the actor $\pi$, determined by the parameters $\theta^\pi$ and one fully connected neural network for the critic $Q$, determined by the parameters $\theta^Q$. The input to both networks is the concatenation of the low-level state $s$ and the low-level subgoal $g_{sub}$. The optimization criterion for the actor $\pi$ is to minimize the $q$ value provided by the critic. The optimization criterion for the critic is to minimize the mean squared error between the critic's output $q$ and the discounted expected reward according to the Bellmann equation for deterministic policies, as described by \autoref{eq:critic_minimi}.
\begin{equation}
\label{eq:critic_minimi}
\begin{aligned}
    & \arg\min\limits_{\theta^{Q}} (q - Q(s_{t},u_{t},g_{sub}|\theta^Q))^2, \\
    \text{where} \quad  
    Q(s_{t},u_{t},g_{sub}|\theta^Q) &= r_t + \gamma \cdot Q(s_{t+1},\pi(s_{t+1},g_{sub}|\theta^\pi),g_{sub}|\theta^Q)
\end{aligned}
\end{equation}

Given that the action space is continuous $n$-dimensional, the observation space is continuous $m$-dimensional, and the goal space is continuous $k$-dimensional with $k \leq m$, the following holds for our theoretical framework: At each step $t$, the agent executes an action $u_t \in \mathbb{R}^n$ given a state $s_t \in \mathbb{R}^m$ and a goal $g \subseteq \mathbb{R}^k$, according to a \new{\textit{behavioral policy}, a noisy version of the \textit{target policy} } $\pi$ that 
%maps from the current observation and goal to a probability distribution over actions. 
deterministically maps the observation and goal to the action\footnote{\new{Applying action noise for exploration purposes is a common practice for off-policy reinforcement learning (e.g., $\epsilon$-greedy). The kind of noise that we investigate in this article (e.g., in the experiments section \autoref{sec:experiments}) is supposed to simulate perceptual errors and not to be confused with this action noise for exploration.}}.
The action generates a reward $r_t=0$ if the goal is achieved at time $t$. Otherwise, the reward is $r_t=-1$. 
To decide whether a goal has been achieved, a function $f(s_t)$ is defined that maps the observation space to the goal space, and the goal is considered to be achieved if $|f(s_t) - g| < \epsilon$ for a small threshold $\epsilon$. This sharp distinction of whether or not a goal has been achieved based on a distance threshold causes the reward to be sparse and renders shaping the reward with a hand-coded reward function unnecessary.

\subsection{Integration of high-level planning with reinforcement learning}
\label{sec:method:integration}
Our architecture integrates the high-level planner and the low-level reinforcement learner as depicted in~\autoref{fig:architecture}.
The input to our framework is a low-level goal $g$. The sensor data that represents the environment state $s$ is abstracted together with $g$ to a propositional high-level description of the world state $S$ and goal $G$. An action planner based on the planning domain definition language (PDDL) \citep{Mcdermott1998_PDDL} takes these high-level representations as input and computes a high-level plan based on manually defined action definitions (c.f. the Appendix in \autoref{apx:planning} for examples). 
We have implemented the caching of plans to accelerate the runtime performance. 
The high-level subgoal state $G_{sub}$ is the expected successor state of the current state given that the first action of the plan is executed. This successor state is used as a basis to compute the next subgoal. To this end, $G_{sub}$ is processed by the subgoal grounding function $f_{subg}$ (Algorithm \autoref{alg:fsubg}) that generates a subgoal $g_{sub}$ as input to the low-level reinforcement learner.

\section{Experiments}
\label{sec:experiments}
This section describes three experiments, designed for the evaluation of the proposed approach. We refer to the experiments as \emph{block-stacking}, \emph{tool use}, and \emph{ant navigation}. The first two experiments are conducted with a Fetch robot arm, and the latter is adapted from research on continuous reinforcement learning for legged locomotion \citep{Levy2019_Hierarchical}. All experiments are executed in the Mujoco simulation environment~\citep{Todorov2012MuJoCo:Control}.
For all experiments, we use a three-layer fully connected neural network with the rectified linear unit (ReLU) activation function to represent the actor-critic network of the reinforcement learner in both experiments. We choose a learning rate of 0.01, and the networks' weights are updated using a parallelized version of the Adam optimizer~\citep{Kingma2015_Adam}. 
We use a reward discount of $\gamma=1 - 1 / T$, where $T$ is the number of low-level actions per rollout. For the block-stacking, we use 50, 100 and 150 low-level actions for the case of one, two and three blocks respectively. For the tool use experiment, we use 100 low-level actions, and for the ant navigation, we used 900 low-level actions.  

Preliminary hyperoptimization experiments showed that the optimal number of units for each layer of the neural networks for actor and critic of the reinforcement learning depends on the observation space.
Therefore, we implement the network architecture such that the number of units in each layer scales with the size of the observation vector. Specifically, the layers in the actor and critic consist of 12 units per element in the observation vector. For example, for the case of the block-stacking experiment with two blocks, this results in 336 neural units per layer (see \autoref{sec:experiments:block_stacking}).
We apply the same training strategy of HER~\citep{Andrychowicz2017HindsightReplay}, evaluate periodically learned policies during training without action noise. We use a fixed maximum number of epochs and early stopping at between 80\% and \new{95\% } success rate, depending on the task. 

In all experiments, we evaluate the robustness of our approach to perceptual noise. That is, in the following we refer to perceptual noise, and not to the action noise applied during the exploration phase of the RL agent, if not explicitly stated otherwise. 
To evaluate the robustness to perceptual noise, we consider the amount of noise relative to the value range of the \new{state } vector. 
To this end, we approximate the continuous-valued state range, denoted  $rng$, as the difference between the upper and lower quartile of the elements in the history of the last 5000 continuous-valued state vectors that were generated during the rollouts\footnote{We use the upper and lower quartile and not the minimum and maximum of all elements to  eliminate outliers.}. For \new{each action step in the rollout } we randomly sample noise, denoted \new{$s_{\gamma}$}, according to a normal distribution with $rng$ being the standard deviation. \new{We add this noise to the state vector. } To parameterize the amount of noise added \new{to the state}, we define a noise-to-signal-ratio $\kappa$ such that the noise added to the state vector is computed as \new{$s_{noisy} = s + \kappa \cdot s_{\gamma}$}. We refer to the noise level as the percentage corresponding to $\kappa$. That is, e.g., $\kappa=0.01$ is equivalent to a noise level of 1\%. 

%For the block-stacking and the tool use experiments, 
\new{For all experiments, i.e., block-stacking, tool use and ant navigation, }
we trained the agent on multiple CPUs in parallel and averaged the neural network weights of all CPU instances after each epoch, as described by \citet{Andrychowicz2017HindsightReplay}. 
\new{Specifically, we used 15 CPUs for the tool use and the block-stacking experiments with one and two blocks; we used 25 CPUs for the block-stacking with 3 blocks. 
For the ant navigation, we used 15 CPUs when investigating the robustness to noise (\autoref{fig:ant-success-rate}) and 1 CPU when comparing our approach to the framework by  \citet{Levy2019_Hierarchical} (\autoref{fig:ant-success-rate-compared}). The latter was necessary to enable a fair comparison between the approaches because the implementation of the architecture of \citet{Levy2019_Hierarchical} supports only a single CPU.  }
For all experiments, an epoch consists of 100 training rollouts per CPU, followed by training the neural networks for actor and critic with 15 batches after each epoch, using a batch size of 256. 
%For the block-stacking experiments we used 25 CPUs, and for the tool-use problem we used 15 CPUs. 
The results in \autoref{fig:results_tower_noise} and  \autoref{fig:rake-success-rate} illustrate the median and the upper and lower quartile over multiple ($n \geq 5$) repetitions of each experiment. 

\subsection{Block-stacking}
\label{sec:experiments:block_stacking}
\autoref{fig:tool_use} (left) presents the simulated environment for this experiment, where a number of blocks (i.e., up to three) are placed randomly on the table. The task of the robot is to learn how to reach, grasp and stack those blocks one-by-one to their corresponding random target location. The task is considered completed when the robot successfully places the last block on top of the others \new{in the right order, } and moves its gripper to another random target location. The difficulty of this task increases with the number of blocks to stack. 
The order in which the blocks need to be stacked is randomized for each rollout. \new{The causal dependencies involved here are, that a block can only be grasped if the gripper is empty, a block (e.g., A) can only be placed on top of another block (e.g., B) if there is no other block (e.g., C) already on top of either A or B, etc.} 

The size of the goal space depends on the number of blocks. For this experiment, the goal space is a subset of the state-space that is constituted by the three Cartesian coordinates of the robot's gripper and three coordinates for each block.
That is, the dimension of the goal- and subgoal space is $k=(1+n_o) \cdot 3 $, where $n_o \in \{1,2,3\}$ is the number of objects.

The state-space of the reinforcement learning agent consists of the Cartesian location and velocity of the robot's gripper, the gripper's opening angle, and the Cartesian location, rotation, velocity, and rotational velocity of each object. That is, the size of the state vector is $|s| = 4 + n_o \cdot 12$, where $n_o$ is the number of blocks. 

The planning domain descriptions for all environments are implemented with the PDDL actions and predicates provided in the Appendix (in \autoref{apx:planning:tower}).

\subsection{Tool use}
\label{sec:experiments:tool}
The environment utilized for this experiment is shown in~\autoref{fig:tool_use} (right). A single block is placed randomly on the table, such that it is outside the reachable region of the Fetch robot. The robot has to move the block to a target position \new{(which is randomized for every rollout) } within the reachable (dark brown) region on the table surface. In order to do so, the robot has to learn how to use the provided rake. The robot can drag the block either with the left or the right edge of the rake. 
The observation space consists of the Cartesian velocities, rotations, and locations of the robot's gripper, the tip \new{of }the rake, and the object. An additional approximation of the end of the rake is added in this task. The goal space only contains the Cartesian coordinates of the robot's gripper, the tip of the rake, and the object.
%The planning domain description for this tool use environment is implemented with the PDDL actions and predicates provided in the Appendix (in \autoref{apx:planning:hook}).
\new{The planning domain description for this tool use environment can be found in the Appendix (in \autoref{apx:planning:hook}).}

\new{
\subsection{Ant navigation}
% \PN{as discussed before, for the motivation of the paper, I highly recommend to move this subsection into the \autoref{sec:result:ant}}
\begin{figure}
    \centering
    \includegraphics[trim=100px 250px 600px 50px,clip,width=0.7\textwidth]{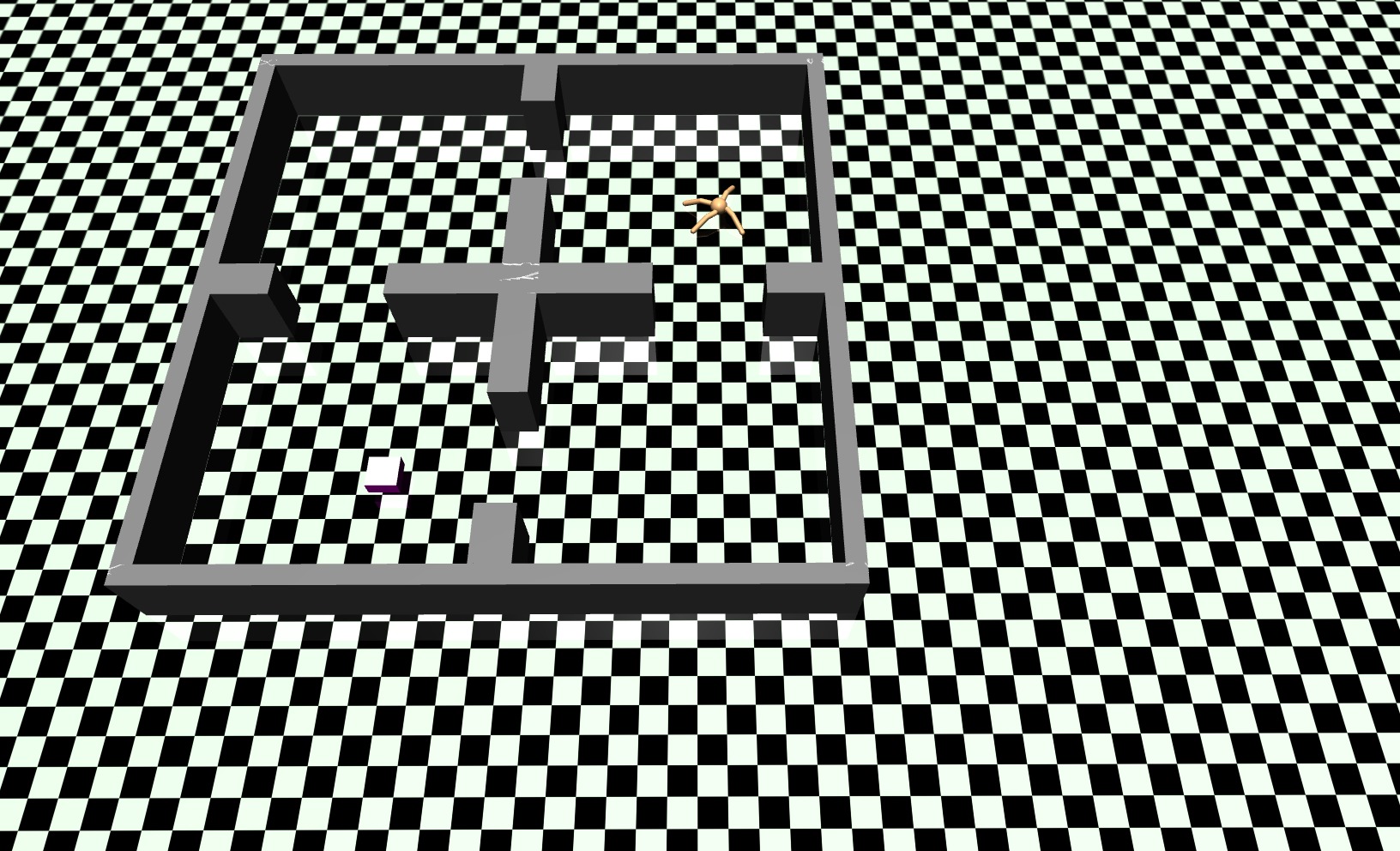}
    \caption{An ant agent performing a navigation and locomotion task in a four-room environment. Herein, the agent needs to learn how to walk and find a way to reach the desired position. In this case, the agent needs to walk from the upper right room to the target location in the lower left room.
     }
    \label{fig:ant_navi}
\end{figure}

 The environment of navigation and locomotion in a four-connected-room scenario is shown in \autoref{fig:ant_navi}, where the ant has to find a way to the randomly allocated goal location inside one of the four-rooms. The state-space consists of the Cartesian location and transitional velocity of the ant's torso, along with the joint position and velocity of the eight joints of the ant (i.e., each leg has one ankle joint and one hip joint). The goal space contains the Cartesian coordinate of the ant's torso. There are no other objects involved in the task.
   \new{The planning domain description and the high-level action specifications for this navigation environment can be found in the Appendix (in \autoref{apx:planning:ant}).}
}

\section{Results}
\label{sec:results}
To evaluate our approach, we investigate the success rate of the testing phase over time for all experiments, given varying noise levels $\kappa$ (see \autoref{sec:experiments}). The success rate is computed per epoch, by averaging over the number of successful rollouts per total rollouts over ten problem instances per CPU. 

\subsection{Block-stacking}
\begin{figure}
    \centering
    \includegraphics[width=\textwidth]{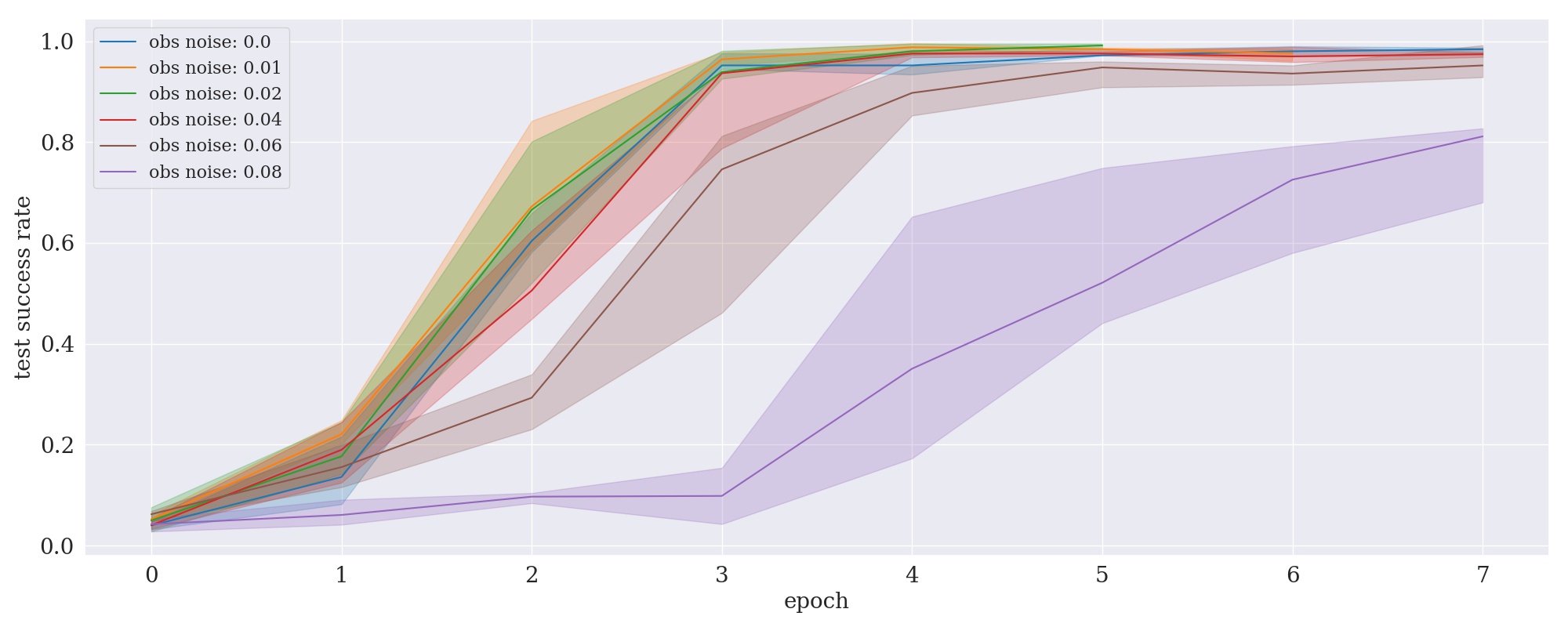}
    \hspace{-3cm}\begin{minipage}[b][86pt][t]{2.4cm}\tiny \color{black!80}{One block} \end{minipage} 
    \includegraphics[width=\textwidth]{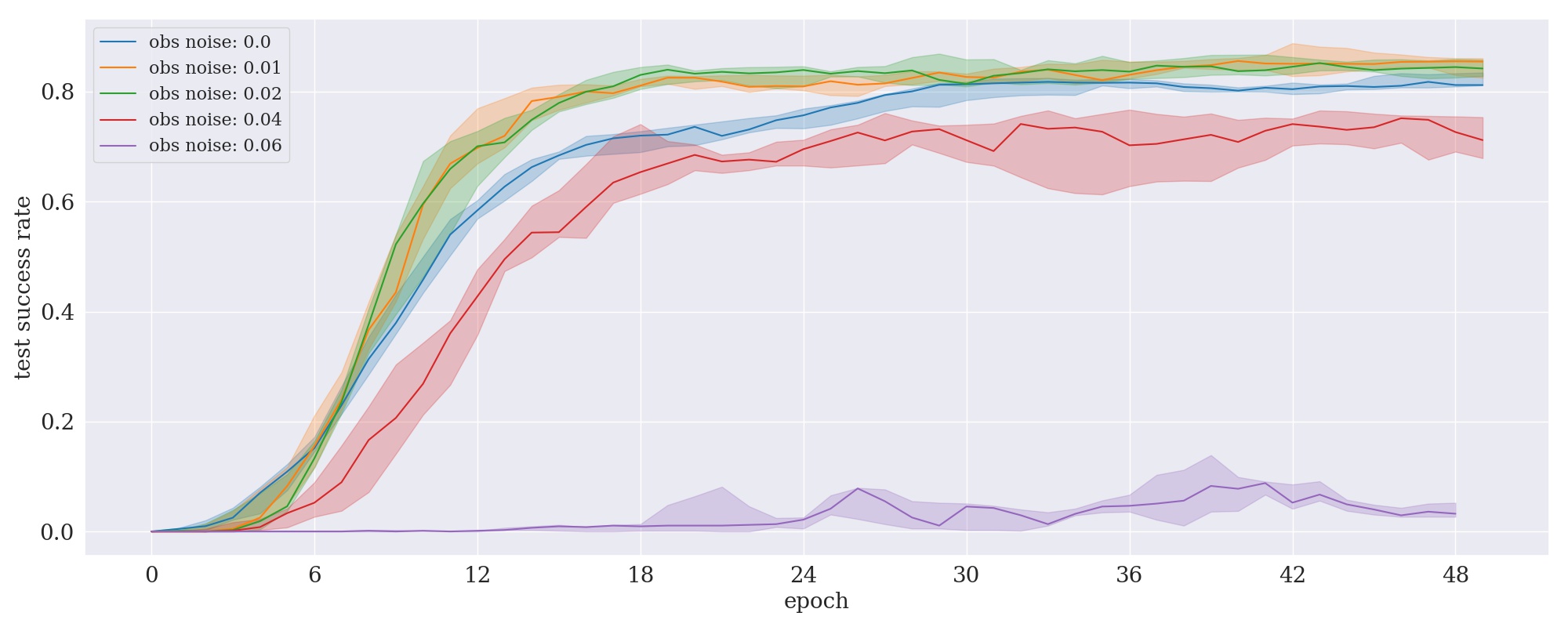}
    \hspace{-3cm}\begin{minipage}[b][86pt][t]{2.4cm}\tiny \color{black!80}{Two blocks} \end{minipage} 
    \includegraphics[width=\textwidth]{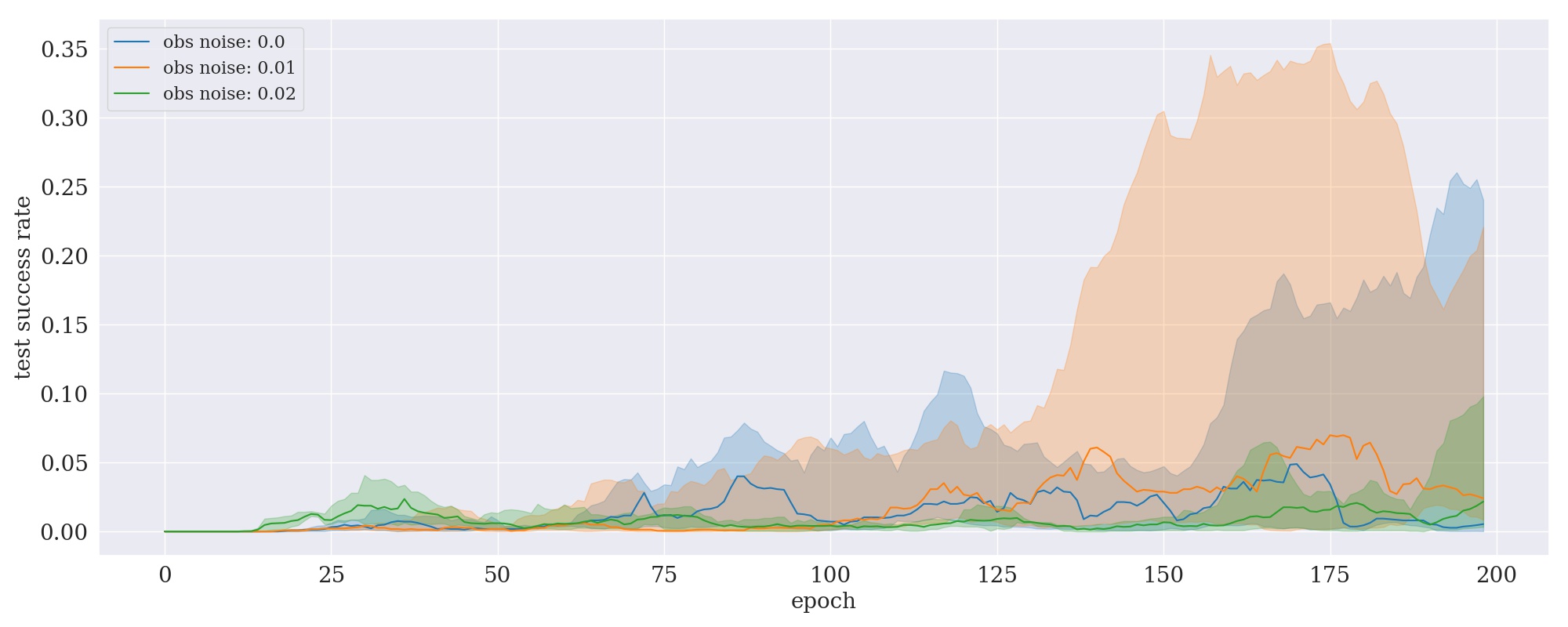}
    \hspace{-3cm}\begin{minipage}[b][86pt][t]{2.4cm}\tiny \color{black!80}{Three blocks} \end{minipage}     
    \caption{Results of the block-stacking experiments for one (top), two (middle) and three (bottom) blocks under different sensory noise levels.}
    \label{fig:results_tower_noise}
\end{figure}

For the experiment with one block, the approach converges after around ten epochs, even with $\kappa = 0.06$, i.e., if up to 6\% noise is added to the observations. The results are significantly worse for 8\% and more noise. For the experiment with two blocks, the performance drops already for 6\% noise. Interestingly, for both one and two blocks, the convergence is slightly faster if a small amount (1-2\%) of noise is added, compared to no noise at all. \new{The same seems to hold for three blocks, although no clear statement  can be made because the variance is significantly higher for this task.}

\begin{figure}
    \centering
    \includegraphics[width=\textwidth]{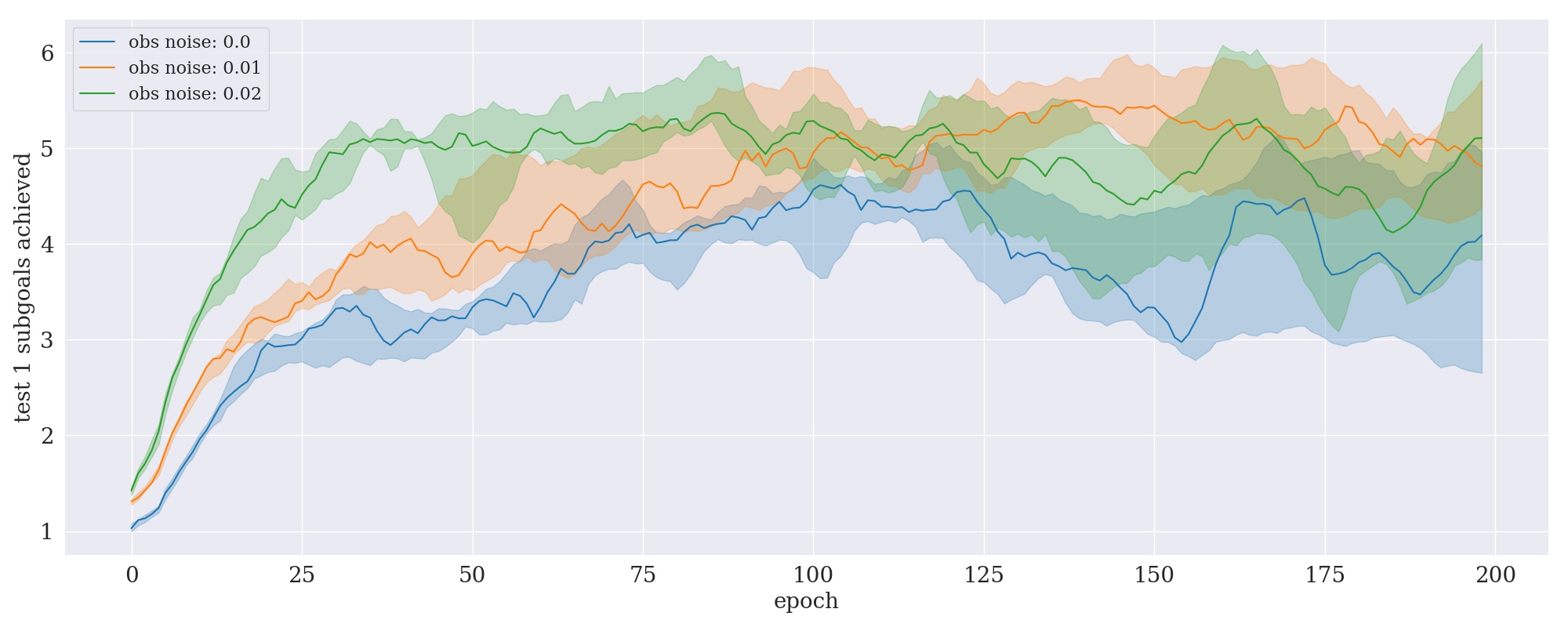}
    \hspace{-3.8cm}\begin{minipage}[b][66pt][t]{2.8cm}\tiny \color{black!80}{Achieved subgoals, three blocks} \end{minipage} 
    \caption{Number of subgoals reached for the case of stacking three blocks.}
    \label{fig:results_tower_noise_subgoals_reached}
\end{figure}

\begin{sloppypar}
For the  case of learning to stack three blocks consider also \autoref{fig:results_tower_noise_subgoals_reached}, which shows how many subgoals have been achieved on average during each epoch. For our particular PDDL implementation, six high-level actions, and hence six sugboals, are at least required to solve the task: \texttt{[move\_gripper\_to(o1), move\_to\_target(o1), move\_gripper\_to(o2), 
    move\_o\_on\_o(o2,o1), move\_gripper\_to(o3), move\_o\_on\_o(o3,o2)]}.
    \new{First, the gripper needs to move to the randomly located object \texttt{o1}, then, since the target location of the stacked tower is also randomly selected, the gripper needs to transport \texttt{o1} to the target location. Then the gripper moves to \texttt{o2} to place it on top of \texttt{o1}, and repeats these steps for \texttt{o3}.}
    The result shows that the agent can consistently learn to achieve the first five subgoals on average, but is not able to proceed further. This demonstrates that the agent robustly learns to stack the first two objects, but fails to stack also the third one.
\end{sloppypar}

\subsection{Tool use}

\begin{figure}
    \centering
    \includegraphics[width=\textwidth]{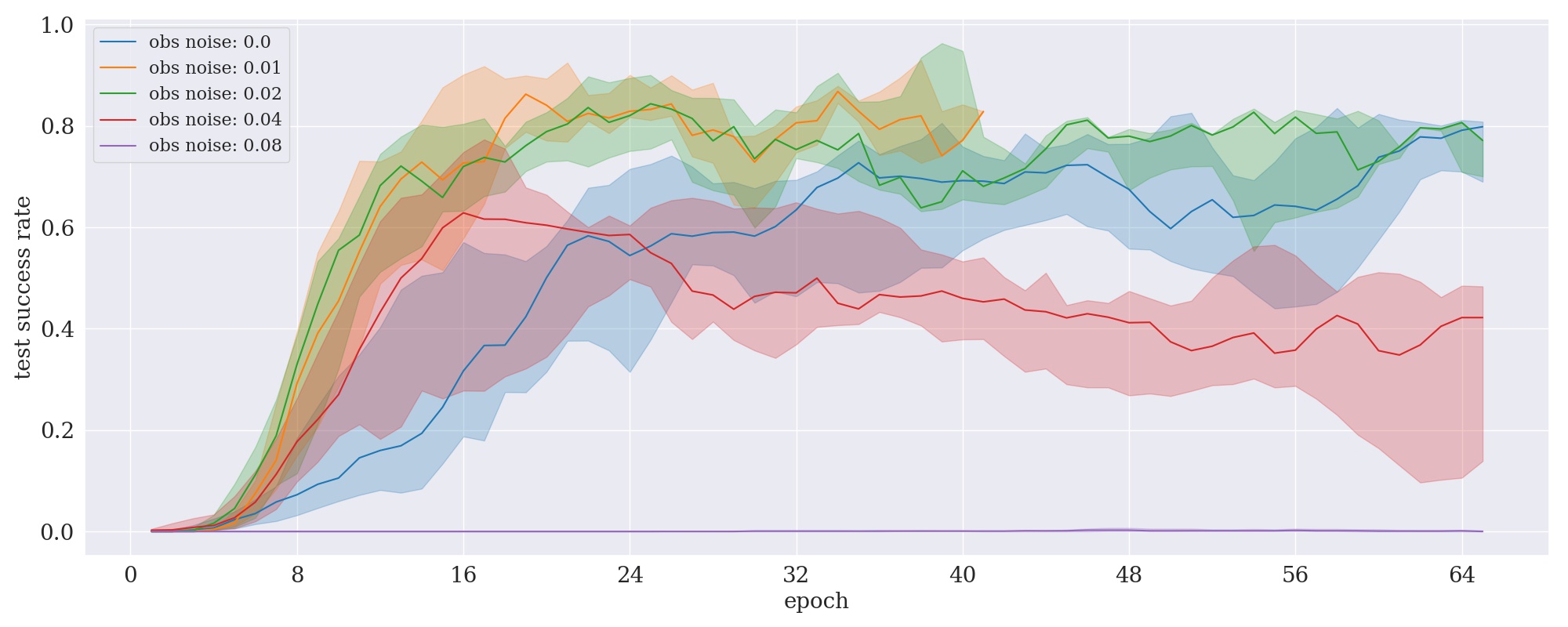}
    \hspace{-3cm}\begin{minipage}[b][66pt][t]{2.4cm}\tiny \color{black!80}{Tool use}\end{minipage} 
    \caption{Results of the tool use experiment under different noise levels. The case for a noise level of 0.01 is subject to the early stopping mechanism.}
    \label{fig:rake-success-rate}
\end{figure}

The results in \autoref{fig:rake-success-rate} reveal that our proposed approach allows the agent to learn and complete the task in under 100 training epochs (corresponding to approximately 8 hours with 15 CPUs) even with a noise level increased up to 4\% of the state range. 
We observe that it becomes harder for the agent to learn when the noise level exceeds 4\%. In the case of 8\% noise, the learning fails 
to achieve a reasonable performance in the considered time--first 100 training epochs (i.e., it only obtains less than 1\% success rate).
Interestingly, in cases with very low noise levels (1\%-2\%), the learning performance is better or at least as good as the case with no noise added at all.

\new{
\subsection{Ant navigation}
\label{sec:result:ant}
\begin{figure}
    \centering
    \includegraphics[width=\textwidth]{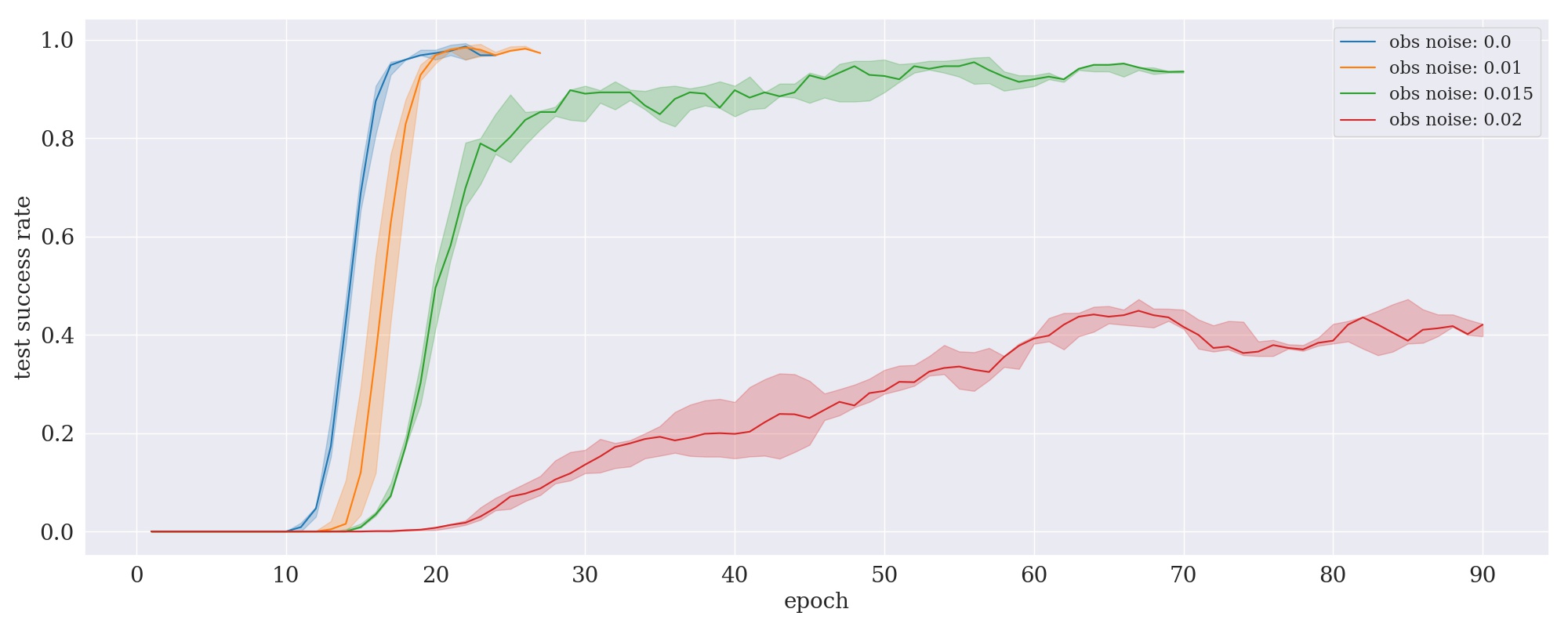}
    \hspace{-3cm}\begin{minipage}[b][66pt][t]{2.4cm}\tiny \color{black!80}{Ant navigation} \end{minipage} 
    \caption{Results of the ant navigation experiment under different noise levels. The curves are subject to early stopping.}
    \label{fig:ant-success-rate}
\end{figure}
\autoref{fig:ant-success-rate} presents the performance of trained agents following our proposed approach in the ant navigation scenario. The results show that the agent can learn to achieve the task in less than 30 training epochs under the low noise level conditions (up to 1\%). The performance decreases slightly in the case of 1.5\% but the agent still can learn the task after around 70 training epochs. With a higher noise level (i.e., 2\%), the agent requires longer training time to cope with the environment.
}

\new{
\subsection{Comparison with hierarchical reinforcement learning}
\begin{figure}
    \centering
    \includegraphics[width=\textwidth]{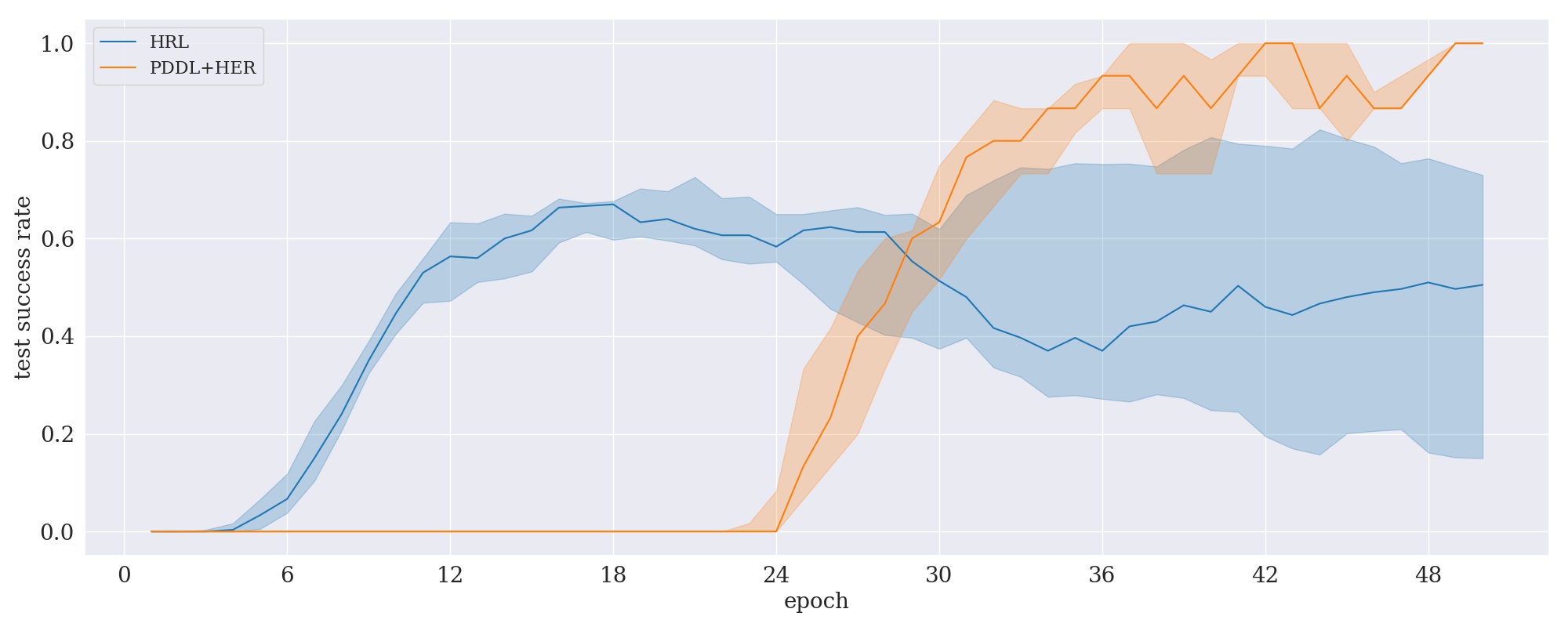}
    \hspace{-5cm}\begin{minipage}[b][66pt][t]{3.4cm}\tiny \color{black!80}{Comparison with HRL, ant navigation}\end{minipage} 
    \caption{Comparison of two approaches for the ant navigation experiment between two approaches: our (PDDL+HER) approach and hierarchical reinforcement learning (HRL)~\citep{Levy2019_Hierarchical}}
    \label{fig:ant-success-rate-compared}
\end{figure}

Results in \autoref{fig:ant-success-rate-compared} depict the benchmark experiment of our proposed approach with the HRL approach by~\cite{Levy2019_Hierarchical}. Though the HRL approach quickly learns the task at the beginning, it does not exceed a success rate of 70 \%. In comparison, our approach learns to solve the task more reliably, eventually reaching 100\%, but the success rate grows significantly later, at around 50 epochs.
}

\section{Discussion}
\label{sec:discussion}

The results indicate that our proof-of-concept addresses the hypotheses \ref{itm:H1} and \ref{itm:H2} as follows:

\subsection{Hypothesis H.1: Ability to ground high-level actions in low-level trajectories}
Our experiments indicate that the grounding of high-level actions in low-level RL-based robot control using the HER approach performs well for small to medium-sized subgoal spaces. However, learning is not completely automated, as the approach requires the manual definition of the planning domain and of the functions $f^p_{subg}$ that maps planning domain predicates to subgoals. 

For the tasks of stacking two blocks and the tool use, the subgoal space involved nine values, and both tasks could be learned successfully. The qualitative evaluation and visual inspection of the agent in the rendered simulation revealed that the grasping of the first and second block failed more often for the experiment of stacking three blocks than for the experiment of stacking two blocks. 
Therefore, we conclude that the subgoal space for stacking three blocks, which involves twelve values, is too large. 

However, the performance on the control-level was very robust. For example, it happened frequently during the training and exploration phase that the random noise in the actions caused a block to slip out of the robot's grip. In these cases, the agent was able to catch the blocks immediately while they were falling down. During the tool use experiment, the agent was also able to consider the rotation of the rake, to grasp the rake at different positions, and to adapt its grip it when it was slipping.

% \subsection{Hypothesis H.2: Ability to solve causal puzzles}
The results indicate that the approach is able to solve causal puzzles if the subgoal space is not too large. The architecture depends strongly on the planning domain representation that needs to be implemented manually. 
In practice, the manual domain-engineering that is required for the planning is appropriate for tasks that are executed frequently, such as adaptive robotic co-working at a production line or in a drone delivery domain. Due to the caching of plans (see \autoref{sec:method:integration}), we have not encountered issues with the computational complexity problem and run-time issues of the planning approach. 

\new{
Our measure of appropriateness that we state in H.1 is to evaluate whether our method outperforms a state-of-the-art HRL approach. \autoref{fig:ant-success-rate-compared} depicts that this is the case in terms of the final success rate. 
Specifically, the figure  shows that the HRL approach learns faster initially, but never reaches a success rate of more than 70\%, while our approach is slower but reaches 100\%. 
A possible explanation for this behavior is that HRL implements a ``curriculum effect'' (cf. \cite{Eppe2019_CGM}), in the sense that it first learns to solve simple subgoals due to its built-in penalization of difficult subgoals. However, as the success rate increases, there are fewer unsuccessful rollouts to be penalized which potentially leads to more difficult subgoals and, consequently, a lower overall success rate. This curriculum effect is not present in our approach because the planning mechanism does not select subgoals according to their difficulty. Investigating and exploiting this issue in detail is potentially subject to further research.
}

%To investigate the performance gain provided by the causal reasoning abilities of the high-level planning layer, we have compared our approach to a \new{non-}hierarchical reinforcement learner (see \autoref{apx:add_experiments}, \autoref{fig:rake-pddl-vs-her-simple}).
%The non-hierarchical approach was not able to learn to solve any of the tasks that we discuss in this paper. Therefore, we investigated a simplified version of the tool use task, which only involves pulling the rake to the target without involving any other objects. For this simplified task, our planning-based approach learned approximately ten times faster than the shallow reinforcement learner. 

\subsection{Hypothesis H.2: Robustness to noise}
\new{For the block-stacking with up to two blocks and the tool-use experiments, the approach converged with a reasonable noise-to-signal ratio of four to six percent. For  the block-stacking with three blocks and for the ant environment, a smaller amount of noise was required. }
\new{An interesting observation is that a very low amount of random noise, i.e., $\kappa=0.01$, improves the learning performance for some cases. Adding random noise, e.g., in the form of dropout, is a common technique to improve neural network-based machine learning because it helps neural networks to generalize better from datasets. One possible explanation for the phenomenon is, therefore, that the noise has the effect of generalizing the input data for the neural network training, such that the parameters become more robust.}

Noise is an important issue for real physical robots. Our results indicate that the algorithm is potentially appropriate for physical robots, at least for the case of grasping and moving a single block to a target location. 
For this case, with a realistic level of noise, the algorithm converged after approximately ten epochs (see \autoref{fig:results_tower_noise}). 
Per epoch and CPU, the agent conducts 100 training rollouts. A single rollout would take around 20 seconds on a physical robot. Considering that we used 15 CPUs, the equivalent robot training time required is approximately 83 hours. 
%This value is reasonable for real applications.
\new{For the real application, } the physical training time can potentially further be lowered by applying more neural network training batches per rollout, and by performing pre-training using the simulation \new{along with continual learning deployment techniques, such as the method proposed by \citet{Traore2019Continual}.}

\section{Conclusion}
\label{sec:conclusion}
We have developed a hierarchical architecture for robotic applications in which agents must perform reasoning over a non-trivial causal chain of actions. We have employed a PDDL planner for the high-level planning and we have integrated it with an off-policy reinforcement learner to enable robust low-level control. 

The innovative novelty of our approach is the combination of action planning with goal-independent reinforcement learning and sparse rewards \citep{Lillicrap2016_DDPG,Andrychowicz2017HindsightReplay}. 
This integration allowed us to address two research problems that involve the grounding of the discrete high-level state and action space in sparse rewards for low-level reinforcement learning. 
We addressed the problem of grounding of symbolic state spaces in continuous-state subgoals (\ref{itm:P1}), by proposing a principled predicate-subgoal mapping, 
\new{which involves the manual definitions of functions $f^p_{subg}$ for each predicate $p$. 

We assume that the manual definition of functions $f^p_{subg}$ generally involves less engineering effort than designing a separate reward function for each predicate. Although this assumption heavily depends on the problem domain and may be subject to further discussion, the manual definition of functions $f^p_{subg}$ is at least a useful scaffold for further research that investigates the automated learning of functions $f^p_{subg}$, possibly building on the research by \citet{Ugur2015}.}

The predicate-subgoal mapping is also required to address the problem of mapping subgoals to low-level action trajectories (\ref{itm:P2}) by means of reinforcement learning with sparse rewards using hindsight experience replay \citep{Andrychowicz2017HindsightReplay}. 
Our resulting approach has two advantages over other methods that combine action planning and reinforcement learning, e.g.,~\citep{Grounds2005_Plan-Q, Yamamoto2018_Learning_AbductivePlanning}: The low-level action space for our robotic application is continuous and it supports a higher dimensionality. 

We have realized and evaluated our architecture in simulation, and we addressed \new{two } hypotheses (\ref{itm:H1} and \ref{itm:H2}): First, we demonstrate that the approach can successfully integrate high-level planning with reinforcement learning 
and this makes it possible to solve simple causal puzzles (\ref{itm:H1}); second, we demonstrate robustness to a realistic level of sensory noise (\ref{itm:H2}). The latter demonstrates that our approach is potentially applicable to real-world robotic applications. \new{The synthetic noise used in our experiments does not yet fully guarantee that our approach is capable of bridging the reality gap, but we consider it a first step towards real robotic applications (see also \cite{Andrychowicz2018_Dexterity,nguyen2018transferring,Traore2019Continual}).}

The causal puzzles that we investigate in this paper are also relevant for hierarchical reinforcement learning (HRL) (e.g., \citep{Levy2019_Hierarchical}), but we have not been able to identify an article that presents good results in problem-solving tasks that have a causal complexity comparable to our experiments. An empirical comparison was, therefore, not directly possible. Our approach has the advantage over HRL that it exploits domain knowledge in the form of planning domain representations. The disadvantage compared to HRL is that the domain knowledge must be hand-engineered. 
In future work, we plan to complement both approaches, e.g., by building on vector-embeddings to learn symbolic planning domain descriptions from scratch by means of the reward signal of the reinforcement learning. A similar approach, based on the clustering of affordances, has been presented by \citet{Ugur2015}, and complementing their method with reinforcement learning suggests significant potential. \new{An overview of this topic and potential approaches is provided by \citet{Lesort2018statel}. }
We also plan to apply the approach to a physical robot and to reduce the amount of physical training time by pre-training the agent in our simulation and by applying domain-randomization techniques \citep{Andrychowicz2018_Dexterity}.

\section*{Conflict of Interest Statement}
The authors declare that the research was conducted in the absence of any commercial or financial relationships that could be construed as a potential conflict of interest.

% \section*{Author Contributions}
% \begin{itemize}[leftmargin=2.5em,labelsep=1em]
%     \item[ME] Manfred Eppe has developed the overall concept and theoretical foundation of the method. He also implemented, developed and evaluated the computational framework together with Phuong D.H. Nguyen. Manfred Eppe has authored the core parts of \autoref{sec:intro} and \autoref{sec:method}, and has contributed significantly to all other sections. 
%     \item[PN] Phuong D.H. Nguyen has researched and investigated significant parts of the related work in \autoref{sec:sota}, and he also worked on further refinements and adaptions of the theoretical background provided in \autoref{sec:method} of this article. He has revised and contributed also to all other sections of this article. Furthermore, Phuong D.H. Nguyen has developed the tool use experiment.
%     \item[SW] Stefan Wermter has actively contributed and revised the article.
% \end{itemize}

% \TODO{@Stefan, what should we write here?}
% The Author Contributions section is mandatory for all articles, including articles by sole authors. If an appropriate statement is not provided on submission, a standard one will be inserted during the production process. The Author Contributions statement must describe the contributions of individual authors referred to by their initials and, in doing so, all authors agree to be accountable for the content of the work. Please see  \href{http://home.frontiersin.org/about/author-guidelines#AuthorandContributors}{here} for full authorship criteria.

\section*{Funding}
% \TODO{fill this}
% The project is supported by the 
Manfred Eppe and Stefan Wermter acknowledge funding by the Experiment! Programme of the Volkswagen Stiftung. Manfred Eppe, Stefan Wermter and Phuong D.H. Nguyen acknowledge support via the German Research Foundation (DFG) within the scope of the IDEAS project of the DFG priority programme ``The Active Self''. Furthermore, we acknowledge supports from the European Union’s Horizon 2020 research and innovation programme under the Marie Sklodowska-Curie grant agreements No 642667 (SECURE) and No 721385 (SOCRATES).

\section*{Acknowledgments}
% \TODO{check this}
We thank Erik Strahl for the technical supports with the computing servers and Fares Abawi for the initial version of the Mujoco simulation for the experiments. We also thank Andrew Levy for providing the code of his hierarchical actor-critic reinforcement learning approach \citep{Levy2019_Hierarchical}.

% \section*{Supplemental Data}
% \TODO{fill this: I don't think we need to add anything here, don't we?}
%  \href{http://home.frontiersin.org/about/author-guidelines#SupplementaryMaterial}{Supplementary Material} should be uploaded separately on submission, if there are Supplementary Figures, please include the caption in the same file as the figure. LaTeX Supplementary Material templates can be found in the Frontiers LaTeX folder.

% \section*{Data Availability Statement}
% The datasets [GENERATED/ANALYZED] for this study can be found in the [NAME OF REPOSITORY] [LINK].
% Please see the availability of data guidelines for more information, at https://www.frontiersin.org/about/author-guidelines#AvailabilityofData

%\bibliographystyle{frontiersinSCNS} % for Science, Engineering and Humanities and Social Sciences articles, for
\bibliographystyle{frontiersinSCNS_ENG_HUMS} % for Science, Engineering and Humanities and Social Sciences articles, for Humanities and Social Sciences articles please include page numbers in the in-text citations
%\bibliographystyle{frontiersinHLTH&FPHY} % for Health, Physics and Mathematics articles
% \bibliography{ideas}
\bibliography{final_bib}

%%% Make sure to upload the bib file along with the tex file and PDF
%%% Please see the test.bib file for some examples of references
% \pagebreak

\section*{Appendix}
\subsection*{PDDL domain descriptions}
\label{apx:planning}
\lstset{
    basicstyle=\ttfamily,
    breaklines=true,
}
\new{For all planning domain definitions, we used only the STRIPS semantic requirement of the Planning Domain Definition Language (PDDL) \citep{Mcdermott1998_PDDL}, i.e., pre- and postconditions, and we realized the quantification operators by grounding the variables manually.}

\renewcommand{\thesubsubsection}{\Alph{subsubsection}}
\subsubsection{Block-stacking}
\label{apx:planning:tower}
The domain description of the block-stacking task is described in the following Listing~\ref{lst:block_dom}. 

\begin{lstlisting}[caption=Block-stacking domain,label={lst:block_dom}]
(define (domain block) 
    (:objects o1 ... on)
    (:predicates 
    	(gripper_at ?o)
    	(gripper_at_target)
    	(at_target ?o)
    	(on ?o1 ?o2)
    )
    (:action move_gripper_to_o
    	:parameters (?o) 
    	:precondition () 
    	:effect (and (gripper_at ?o) (forall ?o1 != ?o: (not (gripper_at ?o1))  (not (on ?o1 ?o)) (not (gripper_at_target))))
    )
    (:action move_o_to_target 
    	:parameters (?o) 
    	:precondition (gripper_at ?o)
    	:effect (at_target ?o)
    )
    (:action move_o_on_o  
    	:parameters (?o1 ?o2) 
    	:precondition (and (gripper_at ?o1) ) 
    	:effect (and (on ?o1 ?o2)   (not (on ?o2 ?o1)))
    )
    (:action move_gripper_to_target 
    	:parameters () 
    	:precondition () 
    	:effect 
    	(and 
    	    (gripper_at_target) 
    	    (forall ?o: (not (gripper_at ?o))
            )
        )
    )
)
\end{lstlisting}
% % https://www.cs.toronto.edu/~sheila/2542/s14/A1/introtopddl2.jpg
% The problem description of the block-stacking task for two blocks and the case where all predicates are initially false is described in the following Listing~\ref{lst:block_prob}. 

% \begin{lstlisting}[caption=Block-stacking problem,label={lst:block_prob}]
% (define (problem pb1) (:domain block)
%     (:init)
%     (:goal (and 
%     	 (at_target o1)
%     	 (gripper_at_target)
%     	 (on o0 o1)
%     ))
% )
% \end{lstlisting}

\subsubsection{Tool use}
\label{apx:planning:hook}
The domain description of the tool use task is described in the following listing~\ref{lst:hook}.

\begin{lstlisting}[caption=Tool use,label={lst:hook}]
(define (domain tool) 
    (:requirements :strips) 
    (:objects obj rake)
    (:predicates 
    	(gripper_at ?o)
    	(gripper_at_target)
    	(at_target ?o)
    	(at ?o1 ?o2)
    )
    (:action move_gripper_to_o 
    	:parameters (?o) 
    	:precondition () 
    	:effect 
    	    (and (gripper_at ?o) 
    	        (forall ?o1 != ?o: 
                    (not (gripper_at ?o1)) 
                    (not (at ?o0 ?o1)) 
                    (not (gripper_at_target))
                )
            )
    )
    (:action move_o_at_o
    	:parameters (?o0 ?o1) 
    	:precondition (gripper_at ?o0)
    	:effect (at ?o0 ?o1)
    )
    (:action move_o_to_target_by_o
    	:parameters (?o1 ?o0) 
    	:precondition (and (at ?o0 ?o1) (gripper_at ?o0) ) 
    	:effect (and (at_target ?o1) (at ?o0 ?o1) )
    )
    (:action move_o_to_target
    	:parameters (?o) 
    	:precondition (gripper_at ?o)
    	:effect (and (at_target ?o) (gripper_at ?o) )
    )
    (:action move_gripper_to_target
    	:parameters (?o)
    	:precondition () 
    	:effect 
    	(and 
    	    (gripper_at_target) 
    	    (forall ?o: 
    	        (not (gripper_at ?o))
            )
        )
    )
)
\end{lstlisting}

\new{
\subsubsection{Ant navigation}
\label{apx:planning:ant}
The domain description of the ant navigation task is described in the following listing~\ref{lst:antnav}.

For this listing we did not use the built-in PDDL objects and variables (indicated with ?$<$objname$>$ syntax) to instantiate the predicates and actions. Instead, we implemented a script to generate the predicate and action definitions according to \autoref{lst:antnav} such that the following criteria are met:

\begin{enumerate}[leftmargin=2.5em,labelsep=1em]
    \item Rooms (denoted $<$R$>$) are labeled 00, 01, 10, and 10, such that the 0 and 1 denote the column and row of the 2x2 grid in the ant navigation environment. E.g., room 00 is the lower left room and room 11 is the upper right room. 
    \item Doors (denoted $<$D$>$, the passages between the rooms) are labeled 0001 0010 0111 and 1011. The labels indicate the passages that connect the rooms. For example door 0001 connects room 00 with room 01. 
    \item For each door $<$D$>$ and room $<$R$>$ we generate the respective predicate names as listed in the \texttt{:predicates} section of the domain definition. 
    \item For each door and room combination we generate the action definitions indicated in the listing below, such that the connections of doors and rooms are appropriate. For example we generate an action definition \texttt{move\_to\_room\_center\_00\_from\_door\_0001} because it is possible to move from door 0001 to the center of room 00. However, we do not generate the action \texttt{move\_to\_room\_center\_00\_from\_door\_0111}, because door 0111 is not connected to room 00.
\end{enumerate}
}
\begin{lstlisting}[caption=Ant navigation,label={lst:antnav}]
(define (domain ant) 
    (:requirements :strips) 
    (:predicates 
    	at_door_<D>          ; whether the agent is at a door
    	at_room_center_<R>   ; whether the agent is at a room-center
    	at_target            ; whether the agent is at the target
    	in_room_<R>          ; whether the agent is inside a room
    	target_in_room_<R>   ; whether the target is inside the room
    )
    
    ; Move from room center of <R> to door <D>
    ; <D1> != <D> is the other door that is adjacent to <R>
    (:action move_to_room_center_<R>_from_door_<D> 
    	:precondition (at_door <D>) 
    	:effect (and 
    	    (at_room_center <R>) 
    	    (not (at_door <D1>)) ) 
    )
    
    ; Move to a door when inside a room that connects to the door
    (:action move_to_door_<D>_from_<R> 
    	:precondition (at_room_center_<R>) 
    	:effect (and (at_door_<D>) (not (at_room_center_<R>)) )
    )
    
    ; Move to the room center of <R> if not at a door <D1> or <D2> of that room
    (:action move_to_room_center_<R> 
    	:precondition (and (in_room_<R>)  (not (at_door_<D1>)) (not (at_door_<D2>))) 
    	:effect (at_room_center_<R>)
    )
    
    ; Move to the target within the room <R>
    (:action move_to_target_in_room_<R> 
    	:precondition (and (at_room_center_<R>) 
    	                   (target_in_room_<R>)) 
    	:effect (and (at_target) (in_room_<R>) 
    	             (not (at_room_center_<R>)) 
    )
)
\end{lstlisting}
\end{document}